\documentclass[11pt]{article}

\PassOptionsToPackage{sort}{natbib}
\usepackage[final]{acl}

\newcommand{\coderepourl}{https://github.com/znhoughton/representation-separability}
\newcommand{\anoncoderepourl}{https://anonymous.4open.science/r/representation-separability-D254/}

\newcommand{\aclmodename}{final}
\newcommand{\codefootnote}{%
  \ifnum\pdfstrcmp{\aclmodename}{review}=0
    \ifdefined\anoncoderepourl
      \footnote{All code and analyses in this paper can be found here: \url{\anoncoderepourl}.}%
    \else
      \footnote{All code and analyses will be released on acceptance; the link is
        omitted here to preserve double-blind review.}%
    \fi
  \else
    \ifdefined\coderepourl
      \footnote{All code and analyses in this paper can be found here: \url{\coderepourl}.}%
    \fi
  \fi}
\newcommand{\weightsfootnote}{%
  \ifnum\pdfstrcmp{\aclmodename}{review}=0
    \ifdefined\anonweightsrepourl
      \footnote{Model weights and checkpoints are released at \url{\anonweightsrepourl}.}%
    \else
      \footnote{A model-weights link is omitted here to preserve double-blind
        review; the weights and checkpoints are released.}%
    \fi
  \else
    \ifdefined\weightsrepourl
      \footnote{Model weights and checkpoints are released at \url{\weightsrepourl}.}%
    \fi
  \fi}

\usepackage{times}
\usepackage{latexsym}
\usepackage[T1]{fontenc}
\usepackage[utf8]{inputenc}
\usepackage{textcomp}   
\usepackage{microtype}
\usepackage{inconsolata}

\usepackage{amsmath}
\usepackage{amssymb}

\usepackage{longtable}
\usepackage{booktabs}
\usepackage{array}     
\usepackage{calc}      
\usepackage{multirow}  
\usepackage{float}     
\usepackage{pdflscape} 

\usepackage{etoolbox}
\newif\ifWasInTwoColumn
\makeatletter
\BeforeBeginEnvironment{longtable}{%
  \if@twocolumn\WasInTwoColumntrue\onecolumn\else\WasInTwoColumnfalse\fi}
\AfterEndEnvironment{longtable}{%
  \ifWasInTwoColumn\twocolumn\fi}
\makeatother

\def\fnum@table{\tablename~\thetable}

\usepackage{graphicx}
\makeatletter
\def\maxwidth{\ifdim\Gin@nat@width>\linewidth\linewidth\else\Gin@nat@width\fi}
\def\maxheight{\ifdim\Gin@nat@height>\textheight\textheight\else\Gin@nat@height\fi}
\def\fps@figure{htbp}
\makeatother

\IfFileExists{upquote.sty}{\usepackage{upquote}}{}  
\usepackage{fancyvrb}

\date{}

\makeatletter
\newsavebox\pandoc@box
\newcommand*\pandocbounded[1]{%
  \sbox\pandoc@box{#1}%
  \Gscale@div\@tempa{\linewidth}{\wd\pandoc@box}%
  \ifdim\@tempa\p@<\p@\scalebox{\@tempa}{\usebox\pandoc@box}\else#1\fi%
}
\makeatother

\title{Everyone Was Right: Lexical Representations in Language Models
Are Both Abstract and Holistic}

\author{
Zachary Nicholas Houghton \\
University of Oregon \\ Vail Systems, Inc. \\
\texttt{znh@uoregon.edu}
\And Yu Zhou \\
Vail Systems, Inc.
\And Dan Pluth \\
Vail Systems, Inc.
\AND Vijay K. Gurbani \\
Vail Systems, Inc.
\And Vsevolod Kapatsinski \\
University of Oregon
}

\begin{document}
\maketitle

\begin{abstract}
Lexical representations can encode both abstract information and
information specific to particular contexts. We investigate how these
forms of information are organized in language models by decomposing
contextual word representations into item, class, and word--class
interaction components. Across six pretrained transformers and two
linguistic categories, we find that the interaction is consistently
present, grows across layers, and contains information associated with
both item and class. Unlike the nearly orthogonal item and class
effects, the interaction overlaps with both and encodes class
information beyond the shared class effect. An interaction also emerges
in artificial-language learners even when the input contains only
additive structure. Together, these findings show that lexical
representations combine abstract information with structure specific to
particular word--class combinations.
\end{abstract}

\section{Introduction}\label{introduction}

Lexical representations are deceptively complex: a single word can carry
vastly different meanings depending on the context, yet these meanings
remain recognizably related. The categorization literature has long
recognized this tension, noting that assigning an instance to a specific
category is rarely straightforward. Simple rule-based approaches quickly
break down: a three-legged cat is undeniably a cat, even if it violates
one of the criteria used to define the category. Ultimately, defining a
lexical category in precise terms may be impossible. Yet, as U.S.
Supreme Court Justice Potter Stewart famously observed, we may not be
able to define a category precisely, but we nevertheless ``know it when
{[}we{]} see it.''\footnote{In \emph{Jacobellis v. Ohio} (1964), Justice
  Stewart famously addressed the difficulty of defining obscenity,
  remarking that while he may not be able to formulate a precise
  definition, he could still recognize obscene material when he saw it.}

This central tradeoff between the context-specificity and the
generalizability of linguistic representations has fueled long-standing
debates over the nature of lexical representations. A substantial body
of work suggests that linguistic representations contain both abstract,
generalizable information as well as information specific to particular
experiences
\citep{morganModelingIdiosyncraticPreferences2015, pierrehumbert2016phonological, kapatsinski2010whatisit}.
For example, children can generalize a productive morphological pattern
to forms they have never encountered before
\citep{berkoChildsLearningEnglish1958}, while lexical representations
can preserve fine-grained information about the contexts in which they
were experienced
\citep{goldinger1998echoes, palmeri1993episodicencoding, bybeeScheibman1999constituency}.
Thus, there is little controversy that linguistic knowledge can contain
both generalizable and item-specific information. The more fundamental
questions for the present study are what gives rise to item-specific
representations, and whether they can emerge even for forms whose
properties are fully predictable from their constituent parts.

\subsection{Item-Specificity in Linguistic
Representations}\label{item-specificity-in-linguistic-representations}

Over the last few decades, a great deal of evidence has accumulated
showing that linguistic representations are not as decomposable into
abstract parts as was traditionally thought. Much of the evidence for
this comes from the phonetics literature, showing that phonetic
representations are surprisingly rich
\citep{goldinger1998echoes, palmeri1993episodicencoding}, as well as the
psycholinguistic literature's line of research demonstrating the
holistic storage of multi-morphemic and multi-word phrases
\citep{bybeeScheibman1999constituency, houghton2025roleabstractrepresentations, houghton2026holisticstorageverb+up, houghton2023doespredictabilitydrive, stembergerAreInflectedForms2004, stembergerFrequencyLexicalStorage1986}.

The strongest evidence for context-specific storage comes from frequency
in favorable context effects
\citetext{\citealp[FFC,][]{raymondBrown2012frequency}; \citealp{bybee2002wordfrequency}; \citealp{brown2004reductionsyllableinitialspanish}; \citealp[see][
for a review]{kapatsinski2026lexicalfrequency}}, in which a form's
behavior is predicted by its frequency in contexts that condition a
change rather than by its overall frequency. FFC effects show that it is
not raw frequency, but rather frequency in a specific context, that is
stored. For example, \citet{raymondBrownHealy2016cumulative} found that
words ending in /t/ or /d/ were more likely to be deleted if they were
words that frequently occurred before a consonant, the conditioning
context for that deletion, regardless of the word's overall frequency.
Crucially, this deletion was observed even when the word occurred
outside the conditioning context \citep{bybee2002wordfrequency}. While
we consider FFC effects to be the clearest evidence for context-specific
storage, they are by no means the only evidence: context-specific
storage is also evidenced by word-specific patterns of phonetic
reduction \citep{bybeeScheibman1999constituency}, the persistence of
subphonemic phonetic detail in memory
\citep{mcmurrayGradientSensitivityWithincategory2008, palmeri1993episodicencoding, miller1994internalstructure},
phrase-specific influences on reading times
\citep{siyanova-chanturia2011seeingphrasetime}, lexically specific
word-order preferences
\citep{morgan2016abstractknowledgedirect, morganModelingIdiosyncraticPreferences2015, morgan2024productiveknowledgeitemspecific},
and phrasal frequency effects on the recognition of their component
units
\citep{kapatsinskiRadicke2009prefabs, houghton2025multiwordrepresentationsminds}.

These findings establish that linguistic representations can encode
information specific to particular words and contexts, but they do not
tell us how widespread such information is. Consider \emph{walk}, which
can function as both a noun and a verb. If its representation as a verb
was fully determined by a representation of \emph{walk} and a
representation of the verb category, then nothing in its representation
would be specific to the \emph{walk}-verb combination. If, instead,
\emph{walk} as a verb had a representation of its own, that
representation would contain information that cannot be reduced to its
constituent parts.

\subsection{Storage Without
Idiosyncrasy}\label{storage-without-idiosyncrasy}

There is little doubt that idiosyncratic words or phrases, such as
idioms, must be stored in some form, since their properties are not
reducible to those of their individual component parts. However, it is
less clear to what extent forms that could in principle be represented
compositionally (i.e., fully regular forms) are, in fact, represented
compositionally.

Usage-based theories have argued that even regular forms are stored
holistically as a function of usage. Bybee \citep{bybee2001} and later
Goldberg \citep{goldberg2019explain} have proposed that frequent words
and phrases can be stored even when their properties are fully
predictable. From this view, storage need not be a consequence of
idiosyncrasy. In fact, it has even been argued that storage is a
prerequisite for a representation to become idiosyncratic in the first
place \citep{bybee2001, goldberg2019explain}.

These debates about the nature of lexical representations often center
around two competing explanations. On one hand, if item-specific
representations arise primarily because linguistic forms have properties
that cannot be predicted from their component parts, then regular forms
should not necessarily have representations that contain information
irreducible to those parts. On the other hand, if frequent forms are
stored with information specific to their use, item-specific
representations may emerge even when a form's behavior is fully
predictable from its component parts.

\subsection{Present study}\label{present-study}

The present study asks how item-specific information is represented and
what properties of the input give rise to item-specific representations.
We focus on representations of words in grammatical classes and
decompose a word-in-class representation into three components: an item
component (e.g., \emph{walk} regardless of the part of speech), a class
component (e.g., a verb component), and an idiosyncratic component,
which captures information specific to the item-in-class combination
(e.g., \emph{walk} as a verb). A central challenge is that a learner's
predictions can conceal the underlying representations: pure exemplar
models can look, from their predictions alone, like they have learned
abstractions \citep{houghton2026exemplars}. Representations, in
contrast, can be examined directly
\citep{pluthInterpretabilityWhisperEncodings2026, hinton1986distributed, elhage2022superposition},
and in a model the input can be controlled as well, so we can ask which
properties of it produce item-specific structure. In particular, do such
representations emerge only when a form has item-specific properties, or
can they also arise for forms whose properties are fully predictable
from their constituent parts? We first validate this decomposition on an
artificial language in which we independently control the amount of
information associated with the item, the class, and their interaction,
then apply it to six pretrained language models across two linguistic
constructions.

Our main contributions are as follows:

\begin{itemize}
\item
  We introduce a method for decomposing word-in-context representations
  into item, class, and idiosyncratic components, and validate it on an
  artificial language.
\item
  We show that MLPs trained on an artificial language and pretrained
  LLMs both learn representations that contain abstract, generalizable
  parts alongside item-specific parts.
\item
  We further show that this learned item-specific knowledge is not
  represented orthogonally to class or item knowledge.
\end{itemize}

\section{Experiment 1: Artificial Languages}\label{sec-exp1}

In Experiment 1, we test our measurement on an artificial language whose
features we control. This allows us to validate our measurement while
simultaneously examining how different features of the language affect a
model's representations.

\subsection{Dataset}\label{sec-exp1-data}

To examine the separability of linguistic representations, we
constructed an artificial language defined by a probability distribution
over a vocabulary of 600 tokens. Each observation combines a
\textbf{form} (one of 60), a \textbf{class} (one of 4), and a
\textbf{context} (drawn from a pool of 400). Each form occurs 24 times
in each class, each time in a different context.

Four effects determine the next-token distribution for a given form,
class, and context. Consider \emph{walk} used as a verb. The
\textbf{item effect} captures information shared by \emph{walk} across
its uses as a noun and a verb, while the \textbf{class effect} captures
information shared by \emph{walk} and other verbs. The
\textbf{interaction effect} captures what is specific to \emph{walk} as
a verb, beyond these two shared effects. Finally, the \textbf{context
effect} captures information associated with the particular context in
which \emph{walk} occurs. Context is drawn independently of form and
class, so it acts as pure nuisance: any interaction the measure recovers
cannot originate in it.

The four effects are combined to define the ground-truth distribution
over next tokens:

\begin{equation}\phantomsection\label{eq-toy}{
\begin{aligned}
&P(\text{next token} \mid f, c, x) = \\
&\quad \operatorname{softmax}\Big( s\big[\, w_\alpha\,\text{item}(f)
  + w_\beta\,\text{class}(c) \\
&\qquad + w_\gamma\,\text{int}(f,c)
  + w_{\text{ctx}}\,\text{context}(x) \,\big] \Big)
\end{aligned}
}\end{equation}

Each effect is constructed using a fixed random matrix whose columns
define the directions in which that effect can vary. The coefficients
determine how those directions combine to produce the effect's
contribution to the next-token logits. For the interaction effect, each
form--class pairing has its own randomly generated vector, independent
of the item and class effects. These vectors are centered across forms
and classes, ensuring that the interaction contains no additive item or
class component. Before the effects are weighted and combined, each is
scaled to unit variance across observations. The weights in
Equation~\ref{eq-toy} therefore control the relative contributions of
the four effects, rather than differences in their initial scale. The
constant \(s = 3\) scales the combined logits before the softmax,
controlling how peaked the resulting probability distribution is.

We vary the item, class, and interaction weights, \(w_\alpha\),
\(w_\beta\), and \(w_\gamma\), factorially over \(\{0, 0.5, 1, 2\}\). A
weight of zero removes the corresponding effect from the language, while
the other values vary its strength relative to the other effects. We
independently vary the context weight, \(w_{\text{ctx}}\), over
\(\{0.5, 2.0, 6.0\}\), ranging from a relatively minor to a dominant
contribution. We then rescale all four weights together to unit length,
such that
\(w_\alpha^2 + w_\beta^2 + w_\gamma^2 + w_{\text{ctx}}^2 = 1\). This
makes the resulting languages more comparable by enforcing the same
total signal (i.e., the variance of the weighted sum in
Equation~\ref{eq-toy} is held constant), despite individual differences
in the amount that each individual term contributes.

We also manipulate the dimensionality of the effects. The item and class
effects have fixed ranks of 8 and 4, respectively, while the interaction
effect takes one of four ranks: \(\{4, 16, 64, 177\}\). The rank
determines the number of independent directions available to represent
an effect. The maximum interaction rank, 177, is the dimensionality
remaining after centering across 60 forms and four classes:
\((60-1)(4-1) = 177\).

For each resulting language, we train models with hidden widths of 8,
16, 32, 128, 256, using either a linear or a ReLU activation. Models are
trained with the Adam optimizer at a learning rate of 0.01, with early
stopping.\footnote{A run counts as converged when the cross-entropy
  fails to improve by more than \(10^{-5}\) over 40 consecutive
  iterations; runs still improving after 3,000 iterations are recorded
  as not converged. Stopping never depends on anything derived from the
  measurement, so it cannot bias the components we report.} We repeat
the full experimental grid with five random seeds, yielding 37,800 runs
in total, of which 99.6\% trained to convergence. Full training details
are provided in Appendix \ref{sec-appendix-grid}.

\subsection{Method}\label{sec-exp1-method}

\subsubsection{The learner}\label{the-learner}

\begin{figure*}

\centering{

\includegraphics[width=1\linewidth,height=\textheight,keepaspectratio]{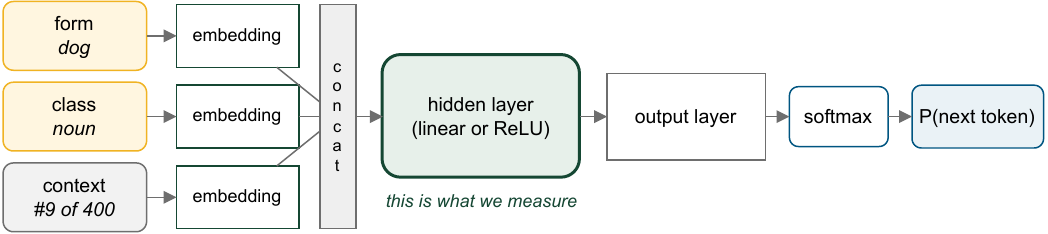}

}

\caption{\label{fig-training}Structure of our model. Every form is
observed 24 times at each class, each time in a different context; that
variation across contexts is what the measurement later splits in two.}

\end{figure*}

The learner model is a multi-layer perceptron (MLP) with a single hidden
layer of width \(d\) and either a linear or a ReLU activation
function.\codefootnote{} The model takes as its input three pieces of
information: form, class, and context. Each is mapped to its own
width-\(d\) embedding, randomly initialized and learned over training,
and the three are concatenated and passed through the model, which then
outputs a probability distribution over the vocabulary
(Figure~\ref{fig-training}). The training objective is thus to minimize
cross-entropy between the model's predicted distribution over the
vocabulary and the exact ground-truth distribution.

\subsubsection{Decomposing a representation}\label{sec-measure}

\begin{figure}

\centering{

\includegraphics[width=0.85\linewidth,height=\textheight,keepaspectratio]{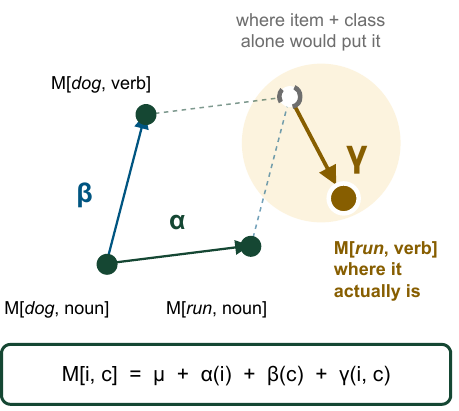}

}

\caption{\label{fig-measurement}The four item-by-class means formed by
two items in two classes. The arrow is \(\gamma(i,c)\).}

\end{figure}

In order to measure the item-specific part of the representation, we
decompose a word-in-context representation into an item component, a
class component, and an idiosyncratic component. In Experiment 1, the
representation we decompose is the MLP's hidden layer representation.
Let \(M[i,c] \in \mathbb{R}^d\) be the mean of those hidden states for
item \(i\) used at class \(c\) (Appendix \ref{sec-appendix-math}). We
decompose those item-by-class means by centering, so that:

\begin{equation}\phantomsection\label{eq-decomp}{
M[i,c] \;=\; \mu + \alpha(i) + \beta(c) + \gamma(i,c),
}\end{equation}

where \(\mu\) is the grand mean; \(\alpha(i)\) is item \(i\)'s mean
across classes minus \(\mu\), which conceptually is the representation
unique to the item; \(\beta(c)\) is the mean across items at class \(c\)
minus \(\mu\), which conceptually is the representation unique to the
class (e.g., noun vs verb); and \(\gamma(i,c)\) is what item \(i\) at
class \(c\) holds that neither marginal accounts for, which conceptually
is the item-specific representation in a specific class
(Figure~\ref{fig-measurement}).

Each of these averages is taken over a sample of observations that
differ only in context, so each carries some of the context term of
Equation~\ref{eq-toy} as noise. We remove it by dividing the
observations into two evenly sized splits, running the decomposition on
each split separately, and estimating each component's size as the dot
product of its two split estimates, which is large when the splits agree
about the component and near zero when they do not. As a result, noise
that differs between the splits is independent and cancels.

For each representation we report two measurements. The first, which we
call the \emph{size} of a component, is the fraction of the
representation accounted for by item, class, and idiosyncratic
information: each component's squared length, summed over its vectors
and divided by the total, so that the three fractions sum to one. The
second, the \emph{overlap} between components, quantifies the extent to
which they occupy the same geometric directions. Specifically, we
project the item vectors onto the class direction and record the
fraction of their squared length that survives, and we project the
interaction onto the smallest subspace containing both the item and the
class vectors (i.e., the span of the two subspaces).\footnote{The class
  effect spans at most one direction fewer than the number of classes,
  so it is low-dimensional in both experiments: one direction in
  Experiment 2 and at most three in Experiment 1. Its projection into
  the item subspace, which has many more dimensions, has high overlap by
  chance; we therefore do not report it.}

To quantify whether a size is meaningful, we redraw the split two
hundred times and estimate how often its confidence interval excludes
zero; for components whose value is exactly zero this flags an effect
about 5\% of the time (Appendix \ref{sec-appendix-validate}), the
false-positive rate expected of a 95\% confidence interval.

To quantify whether an overlap is meaningful, we project the effect,
held exactly as measured, onto two hundred random subspaces of the same
rank, which gives its chance level: the distribution of overlaps that an
unrelated subspace of the same rank would yield. An overlap below its
chance level indicates meaningful orthogonality; an overlap above it
indicates meaningful non-orthogonality.

It's important to note that the weights in Equation~\ref{eq-toy}
describe the ground-truth distribution the model is trained to match,
while the measure decomposes the representation that the model
\emph{learns}, so the two sets of numbers are not necessarily analogous:
a learner handed a language that is 85\% interaction can learn a
representation in which the interaction is far smaller (or larger).
Appendix \ref{sec-appendix-validate} reports a separate check in which
components were set directly in a representation and recovered almost
exactly, which is what licenses interpreting the results below as facts
about the learner rather than about the measure. We include runs only if
at least one of the three components has a re-split interval excluding
zero; where none does, the three shares divide noise.

\subsubsection{Analysis}\label{sec-exp1-analysis}

We model our data using an ordered beta regression
\citep{kubinec2023ordbeta} for the sizes and a Gaussian model for the
overlaps, both in \emph{brms} \citep{burkner2017brms} with weakly
informative priors (See Appendix \ref{sec-appendix-grid} for full
details). For the sizes we fit one model over all three components,
where the outcome variable is the component's share of the
representation and group is that component. Fixed-effects for each
component, the width of the hidden-layer, and total rank of the language
were included along with maximal random effects \citep{barr2013maximal}.
Equation~\ref{eq-model} gives the full model equation. All predictors
are centered. A language here is one combination of the four weights and
the interaction rank; a run here is the five seeds for a given language.

The overlaps are likewise fit as one model over both overlap measures,
with the same structure. Outcome is the log of an overlap divided by its
own chance level, and Group is the overlap measure: the item effect's
overlap with the class direction, or the interaction's overlap with the
span of the item and class subspaces.

\begin{equation}\phantomsection\label{eq-model}{
\begin{aligned}
& \text{Outcome} \sim \text{Group} * \big( \text{ItemEffect} \\ & \quad + \text{ClassEffect} + \text{InteractionEffect} \\ & \quad + \text{log2(Width)} * \text{log2(TotalRank)} \big) \\ & \quad + (1 + \text{log2(Width)} \mid \text{Language}) \\ & \quad + (1 \mid \text{Run})
\end{aligned}
}\end{equation}

\subsection{Results}\label{sec-exp1-results}

Results are visualized in Figure~\ref{fig-toy}, with the full
coefficient results included in Appendix \ref{sec-appendix-capacity}.
Where the hidden layer is wide enough, the learners reproduce the
language closely (median Kullback-Leibler divergence of 0.0016 at the
widest, against 0.6473 at the narrowest; Figure~\ref{fig-toy-fit}
reports the fit across the whole grid).

\begin{figure}

\centering{

\includegraphics[width=1\linewidth,height=\textheight,keepaspectratio]{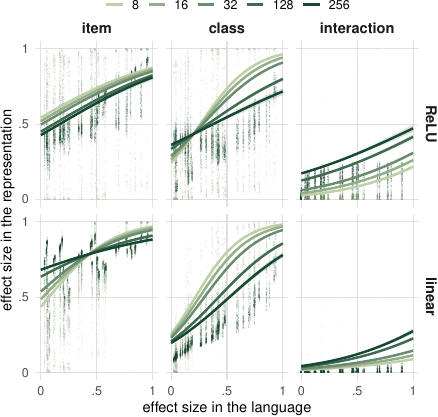}

}

\caption{\label{fig-toy}Ground-truth effect size of each component in
the language (x) against its size in the MLP's representations (y).
Points are runs; lines are fitted values from the size model
(Equation~\ref{eq-model}) with 95\% intervals, colored by hidden width.}

\end{figure}%

Unsurprisingly, the effect size for each component increases as a
function of the size of the effect planted in the language (and since
the three components sum to 1, the other two components definitionally
decrease). Of the three components, the item effect is strongest across
every width (Figure~\ref{fig-toy}).

The size of the interaction barely depends on its rank on its own
(\(\beta = -0.013\)), which is by design: the interaction is scaled to
the same variance at every rank, so raising its rank spreads it over
more directions rather than making it larger. What governs how much of
the interaction is encoded is the width of the hidden layer
(\(\beta = +0.314\)). While rank alone does not affect the size of the
interaction, it does interact with width (\(\beta = +0.017\)): at narrow
widths a higher-rank interaction needs more directions than the layer
has and the excess is lost, while once the layer is wide enough the
model represents more of it. This is not the case for the item effect,
which decreases as the hidden layer grows (\(\beta = -0.090\)), or for
the class effect, which barely changes (\(\beta = +0.028\)). In the
linear condition width is spent on the item effect at the class effect's
expense (\(\beta = +0.123\) against \(\beta = -0.123\)); the interaction
grows with width there too (\(\beta = +0.157\)) but stays small
throughout.

Where no interaction is planted, a ReLU learner still represents one:
the measured value is small (median 0.0056 at the widest layer) but its
interval excludes zero in 45\% of runs, against a nominal 5\%. The
linear model cannot do this at any width, since a linear map of additive
inputs is additive.

The amount of overlap is also affected by the width of the hidden layer:
For the ReLU models, the wider the hidden layer, the less the item and
class effects share directions (\(\beta = -0.383\)). Similarly, the
overlap between the interaction and the item-class subspace follows the
same pattern (\(\beta = -0.301\)). The linear models show a different
pattern, however. The item-class overlap shows no effect as a function
of width, and the interaction effect's overlap with the item-class space
grows as a function of the width (\(\beta = +0.106\)), climbing from
0.45 to 0.78 times chance. It approaches chance without reaching it, and
both overlaps stay below chance at every width under the linear learner
(Appendix \ref{sec-appendix-capacity}). Smaller hidden layers thus do
not merely represent less of the interaction, they also force the
components onto shared directions.

\section{Experiment 2: Language Models}\label{sec-exp2}

Experiment 1 asks what a learner represents when we control the
composition of its language; Appendix \ref{sec-appendix-validate}
separately demonstrates that the measure recovers structure planted
directly in a representation. Neither speaks to how a transformer
represents real language, which is the question Experiment 2 asks.
Experiment 2 applies the same decomposition to the contextual
representations of pretrained language models, across two linguistic
categories: part of speech and grammatical role.

\subsection{Dataset}\label{sec-exp2-data}

In order to examine the separability of linguistic representations in
models, we examined two linguistic categories: part of speech and
grammatical role. For both, we concatenate six general-domain English
Universal Dependencies treebanks \citep{nivre2020ud}, giving 41,095
sentences and 706,323 word tokens; Appendix \ref{sec-appendix-stimuli}
provides the full breakdown. We measure two model families at three
matched scales: OPT-BabyLM, trained on developmentally plausible
quantities of child-available language
\citep{warstadt2023babylm, houghton2026holisticstorageverb+up}, and
Pythia, trained on the Pile \citep{biderman2023pythia}, at 125M/160M,
350M/410M, and 1.3B/1.4B parameters, with widths of 768, 1024, and 2048.
Appendix \ref{sec-appendix-models} gives parameter counts and
training-token estimates.

\subsection{Stimuli}\label{sec-exp2-stimuli}

Our stimuli comprise two linguistic categories: part of speech
(\textsc{noun} versus \textsc{verb}) and grammatical role (\emph{nsubj}
versus \emph{obj}, restricted to nouns). For each category we select
items that occur in both of its classes, such as \emph{walk}, which
occurs as both a noun and a verb. An item is a single inflected word
form, not a lemma, so different inflections of one lemma are separate
items: \emph{walk} and \emph{walking} are both items, each included only
if it is itself attested in both classes. Pooling them under one lemma
would collect \emph{walk} and \emph{walking} into a single item, and the
two classes would then be comparing different tokens, which the
embedding layer could then separate before any contextualization has
occurred. Capitalization is ignored when occurrences are grouped, but
not in the text the models read, so a sentence-initial \emph{Walk}
counts toward the item \emph{walk} and is tokenized as it appears
there.\footnote{OPT-BabyLM's tokenizer is uncased, so this choice has no
  effect for that family of models. On the other hand, Pythia's
  tokenizer is not uncased; however, lowercasing the input would be the
  larger distortion, since a sentence-initial word is capitalized in
  written English which the model was trained on.}

An item is included only if it occurs at least ten times in each class.
This yields 78 items for part of speech and 49 for grammatical role,
with median counts per class spanning from 15 to 19 (Appendix
\ref{sec-appendix-stimuli}), inside the range we validated for the
measurement (Appendix \ref{sec-appendix-validate}).

\subsection{Method}\label{sec-exp2-method}

The decomposition, the two reported quantities, and the replication
logic are identical to Experiment 1. For every token we extract the
hidden state of the word's last subword at every layer of each
model.\footnote{Almost every item is a single subword, so this choice
  has little bearing on the results: zero items are split into multiple
  subword tokens under Pythia's tokenizer, and only eight are split
  under OPT-BabyLM's tokenizer.}

Transformer hidden states contain a few rogue dimensions whose means and
variance are both disproportionately high \citep{timkey2021bark}, and
these would otherwise dominate any subspace estimated from the raw
embeddings, so we divide each dimension by its standard deviation before
decomposing (Appendix \ref{sec-appendix-standardize}).

\subsection{Results}\label{sec-exp2-results}

The results are visualized in Figure~\ref{fig-llm}, but further details
can be found in Appendix \ref{sec-appendix-llm}.

The item effect is the largest component in both categories, running
from 0.657 to 0.846 for part of speech and 0.828 to 0.904 for
grammatical role. Similarly, the effect of class is mostly comparable
across the two linguistic constructions (0.039 to 0.078 vs 0.050 to
0.098 for part of speech and grammatical role respectively). On the
other hand, while the interaction effect is larger than the class effect
for part of speech (0.096 to 0.245), the same is not true for
grammatical role (0.045 to 0.094). This may indicate that part of speech
is more lexicalized than grammatical role, perhaps unsurprisingly.

Another way to describe these results is with respect to the whole
representation rather than the item-by-class structure: the item effect
accounts for 19\% to 38\% of what varies in a token, the class effect
1\% to 3\%, and the interaction 2\% to 8\%, with the remaining 56\% to
72\% being specific to the context a word appears in (See Appendix
\ref{sec-appendix-llm}). The interaction is not, however, inherited from
that context: at the deepest layer only a small fraction of it is
attributable to the context the word appears in; the majority of the
interaction is unique to the lexical representation itself (Appendix
\ref{sec-appendix-context}).

Both the class effect and the interaction effect increase as a function
of layer depth. The interaction is essentially absent at layer 0 in both
categories, between 0.0009 to 0.0088 across every category and model,
and rises to between 0.0455 to 0.2449 (Figure~\ref{fig-llm}). This is
not surprising since at the first layer, the model has yet to integrate
enough context to commit to the class of the representation.

Turning to how those components are oriented, the class direction
carries almost none of the item effect. At the deepest layer, item-class
overlap runs 0.0035 to 0.0067 for part of speech and 0.0033 to 0.0062
for grammatical role, so projecting out the class direction leaves more
than 99\% of the item effect intact. While both overlaps are technically
less orthogonal than a random subspace, the overlap itself is tiny, so
the two are near-orthogonal rather than exactly orthogonal.

In contrast to the main effects, the interaction effect is not
orthogonal to class or item. Across the six models, 0.106 to 0.239 of
the interaction for part of speech is in the span of the class and item,
and for grammatical role 0.071 to 0.179 is, against nulls whose upper
ends run 0.023 to 0.058 and 0.014 to 0.038 respectively. In other words,
the small departure from orthogonality between item and class becomes a
large one for the interaction.

The LLMs show a different pattern from the artificial learners: In the
artificial learner, the components became meaningfully orthogonal once
the hidden layer was wide enough, but for LLMs the components remain
meaningfully non-orthogonal at every scale (3.9 to 13.0 times chance for
the item effect into the class direction and 3.5 to 11.2 for the
interaction into the span of both), and increasingly so as width grows.
Additional width is spent representing more of the language rather than
giving the same structure more room.

\begin{figure}

\centering{

\includegraphics[width=1\linewidth,height=\textheight,keepaspectratio]{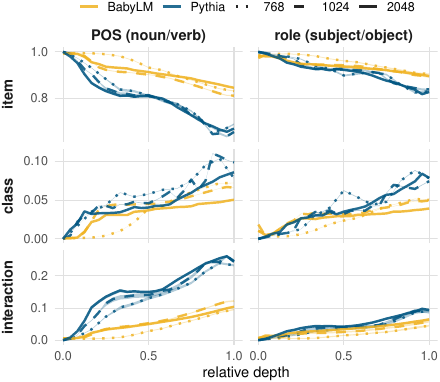}

}

\caption{\label{fig-llm}Fraction of the representation (ignoring
context) accounted for by each component, summing to 1, against relative
depth from the embedding layer (0) to the final layer (1). Each line is
one model; color is the training corpus, linetype the hidden width,
shading the 95\% interval.}

\end{figure}%

\section{Experiment 3: Probing the Interaction}\label{sec-exp3}

Experiment 2 shows that lexical representations contain an idiosyncratic
component that can't be separated out into simple item and class
representations, but neither shows what information is contained within
the interaction. An item-by-class term captures any reliable difference
between a word's uses in the two classes; that difference is not an
artifact of the differing contexts in which the two classes occur
(Appendix \ref{sec-appendix-context}), and here we use a probe to
examine whether the interaction effect \(\gamma\) contains class
information.

\subsection{Method}\label{sec-exp3-method}

We test whether the interaction encodes class information by probing its
representation. If \(\gamma\) encodes the class, the class should be
recoverable from the part of a token's representation that is neither
the item effect nor the shared class direction, and we remove both
explicitly. All probing happens \emph{within} a single word, so we ask
whether \emph{walk}'s noun uses can be told from \emph{walk}'s verb
uses; the item effect is constant within a word and so cannot
contribute. We then estimate \(\beta\) as the difference between the two
class means pooled over all items, normalize it, and remove that
component from every token, so that whatever separates the classes
afterwards is not the direction they share.

What remains after removing those components is the interaction: the
part of a word of a specific class that is neither a fact about the word
nor a fact about the class. We fit a cross-validated logistic probe to
it, five folds, with classes balanced by subsampling within each word so
that chance is 0.50, and average the resulting accuracies over words. We
include tokens only if they have at least ten occurrences in each class,
the same threshold the stimuli are selected on. The whole procedure is
then repeated with the class labels permuted within each word,
preserving each word's observation count and class balance while
disrupting the correspondence between individual observations and their
class labels. This provides a permutation-based control related to that
of \citet{hewitt2019control}, but differs in preserving each word's
class balance rather than assigning each word type a consistent random
label. We run this at the deepest layer of the largest model in each
family, for both categories.

\subsection{Results}\label{sec-exp3-results}

Results are reported in Table~\ref{tbl-decode}. For all models, the
class is recoverable from the interaction at well above chance, reaching
0.94 for part of speech and 0.84 for grammatical role, while the
permuted null sits on chance.

\begin{table}

\centering{

\centering\begingroup\fontsize{7}{9}\selectfont

\begin{tabular}[t]{lll}
\toprule
model & accuracy & null\\
\midrule
\addlinespace[0.3em]
\multicolumn{3}{l}{\textbf{part of speech}}\\
\hspace{1em}BabyLM 1.3B & 0.90 [0.88, 0.91] & 0.51 [0.49, 0.54]\\
\hspace{1em}Pythia 1.4B & 0.94 [0.92, 0.95] & 0.50 [0.48, 0.53]\\
\addlinespace[0.3em]
\multicolumn{3}{l}{\textbf{grammatical role}}\\
\hspace{1em}BabyLM 1.3B & 0.80 [0.77, 0.83] & 0.50 [0.47, 0.53]\\
\hspace{1em}Pythia 1.4B & 0.84 [0.81, 0.86] & 0.52 [0.49, 0.55]\\
\bottomrule
\end{tabular}
\endgroup{}

}

\caption{\label{tbl-decode}Accuracy of recovering the class from the
interaction, at the deepest layer of the largest model in each family,
compared to the null; brackets are bootstrapped intervals.}

\end{table}%

The results demonstrate that \(\gamma\) carries class information.

\section{Discussion and Conclusion}\label{sec-discussion}

The present study examined whether an in-context word's representation
in language models is abstract or item-specific and found that it
contains elements of both. We decomposed a word's representation into an
item effect, a class effect, and a word-class interaction. In a model
trained on an artificial language, we found an interaction effect even
where the language itself was additive, so some of this structure comes
from the architecture rather than from the language. Additionally,
across six language models and two linguistic categories, we found a
similar interaction effect. The abstract components were nearly
orthogonal, whereas the interaction overlapped with both, indicating
that abstract and item-specific information are represented together
rather than on separate axes.

Overall, the present results indicate that language model
representations encode both abstract and item-specific structure,
supporting generalization across words and classes while retaining
information about specific combinations.

\section*{Limitations}\label{limitations}
\addcontentsline{toc}{section}{Limitations}

Our evaluation is restricted to English and relies on two linguistic
categories tested across two model families (OPT-BabyLM and Pythia) at
three parameter scales. While these choices cover a range of
architectures and linguistic phenomena, broader typological,
categorical, and architectural validation remains a direction for future
work.

\bibliography{refs}
\renewcommand{\bibliography}[1]{}
\clearpage

\onecolumn

\appendix

%
%
\raggedbottom
\floatplacement{figure}{H}
\floatplacement{table}{H}

\section{The Artificial-Language Grid}\label{sec-appendix-grid}

Table \ref{tbl-grid} describes the language and the MLP parameters.

\begin{table}

\centering{

\centering\begingroup\fontsize{8}{10}\selectfont

\begin{tabular}[t]{ll}
\toprule
factor & levels\\
\midrule
\addlinespace[0.3em]
\multicolumn{2}{l}{\textbf{language}}\\
\hspace{1em}item weight & 0, 0.5, 1, 2\\
\hspace{1em}class weight & 0, 0.5, 1, 2\\
\hspace{1em}interaction weight & 0, 0.5, 1, 2\\
\hspace{1em}context weight & 0.5, 2, 6\\
\hspace{1em}interaction rank & 4, 16, 64, 177\\
\addlinespace[0.3em]
\multicolumn{2}{l}{\textbf{learner}}\\
\hspace{1em}hidden width & 8, 16, 32, 128, 256\\
\hspace{1em}activation & linear, ReLU\\
\addlinespace[0.3em]
\multicolumn{2}{l}{\textbf{replication}}\\
\hspace{1em}seed & 0, 1, 2, 3, 4\\
\bottomrule
\end{tabular}
\endgroup{}

}

\caption{\label{tbl-grid}The Experiment 1 language parameters. The first
five rows define each artificial language, the next two the learner MLP
model. Item and class ranks are held fixed so that the interaction's
rank is the only one that varies. Crossing every level gives 38,400
cells; the three languages in which all three effect weights are zero
contain no item-by-class structure at all and are excluded, leaving the
reported total.}

\end{table}%

Item rank is fixed at 8 and class rank at 4, so the structure a language
requires is 12 directions plus the interaction's rank, running from 16
to 189. Of the 37,800 runs, 99.6\% trained to convergence.

Every learner is trained in full batch with the Adam optimizer at a
learning rate of 0.01, minimizing cross-entropy against the exact
ground-truth distribution until the loss plateaus, on one GPU with
twelve worker processes.

Both the size and the overlap models use weakly informative priors
rather than the package defaults: a normal distribution with mean zero
and standard deviation 2.5 on every population-level coefficient, a
normal distribution with mean zero and standard deviation 1 on the
standard deviations of the random effects, and an LKJ(2) prior on their
correlation. The coefficient prior still covers the largest effects
either model estimates, which reach about 2.8 on the logit scale, and
the prior on the standard deviations is much wider than any variance
component the data support, so neither prejudges a result.

\begin{figure}

\centering{

\includegraphics[width=1\linewidth,height=\textheight,keepaspectratio]{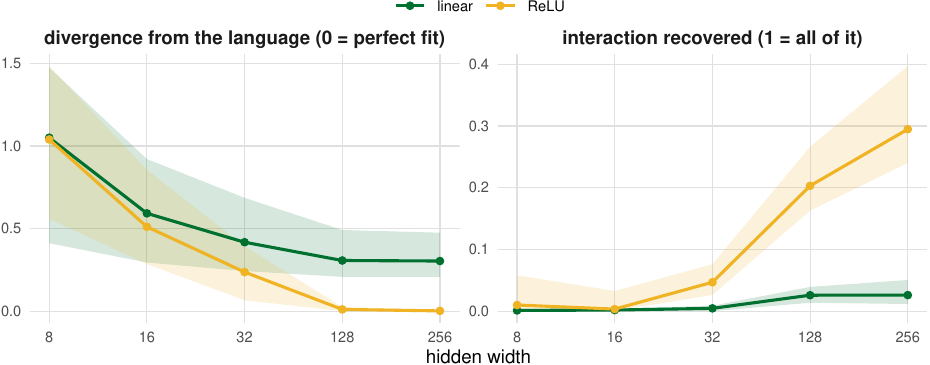}

}

\caption{\label{fig-toy-fit}Fit and recovery against hidden width, for
languages whose planted structure is more than 30\% interaction. Left:
Kullback-Leibler divergence from the learner to the ground-truth
distribution, so zero is a perfect fit. Right: the interaction share
measured in the hidden layer as a fraction of the share planted in the
language, so one would be full recovery and the plotted values are the
fraction of it reached. Lines are medians over runs with interquartile
bands, which show how much the languages in the grid differ from one
another rather than measurement error.}

\end{figure}

Figure~\ref{fig-toy-fit} reports how well that training succeeded, and
what it did not buy. The left panel is the Kullback-Leibler divergence
from the learner's predictions to the ground-truth distribution, which
is the cross-entropy at convergence minus the entropy of the target; the
entropy floor itself varies across the grid, so the raw cross-entropy is
not comparable between languages but this difference is. The right panel
is the share of the item-by-class structure that the hidden layer
actually carries, as a fraction of the share planted in the language.
Both are restricted to the languages in which the interaction is more
than 30\% of the planted structure, which are the languages where the
question has teeth.

\newpage{}

\section{The Decomposition}\label{sec-appendix-math}

\begin{figure}

\centering{

\includegraphics[width=1\linewidth,height=\textheight,keepaspectratio]{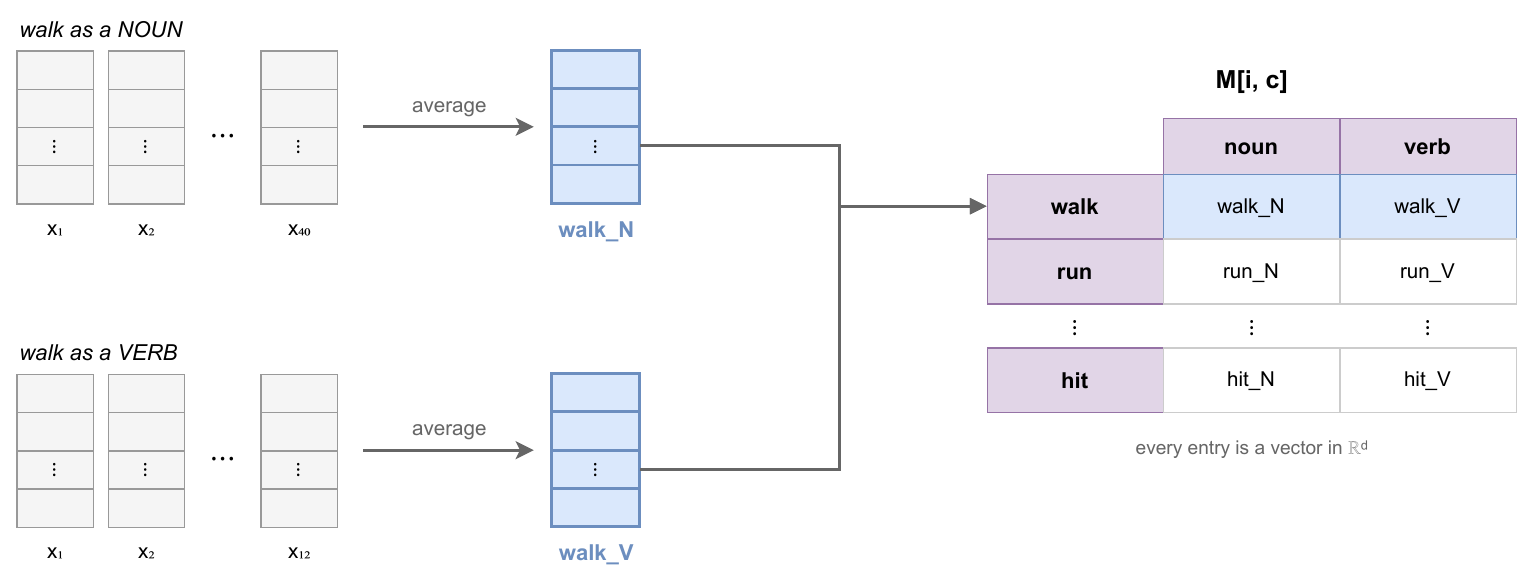}

}

\caption{\label{fig-grid}Each word in each class is observed in many
contexts, and those observations are averaged to give that word's
representation in that class. Every one of these averages is a vector in
\(\mathbb{R}^d\), and every item appears in every class.}

\end{figure}%

\begin{figure}

\centering{

\includegraphics[width=1\linewidth,height=\textheight,keepaspectratio]{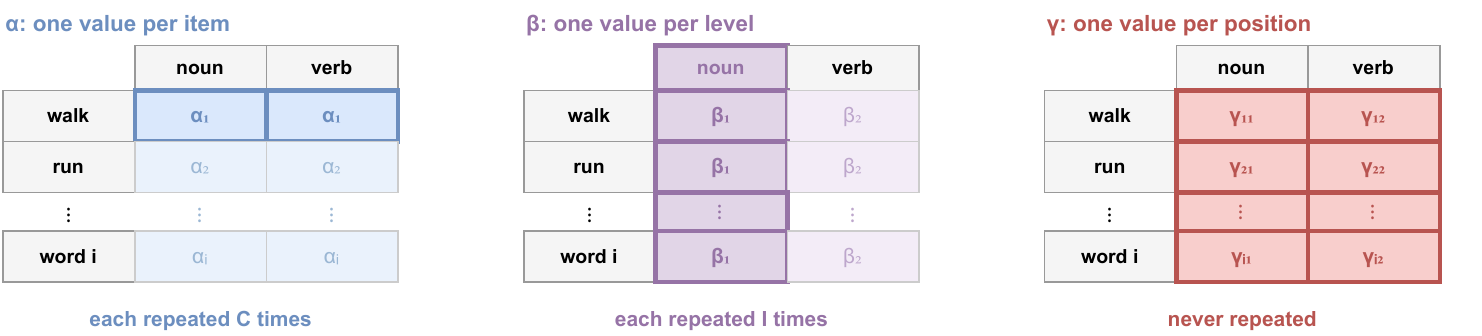}

}

\caption{\label{fig-replication}The three components across the grid of
words and classes. An item effect is one value per word, the same in
every class; a class effect is one value per class, the same for every
word; and the interaction differs for every word in every class.}

\end{figure}

A word used in a given class is observed in many contexts, and the
representation differs a little in each. Averaging those observations
gives that word's representation in that class: for items
\(i \in \{1,\dots,I\}\) crossed with classes \(c \in \{1,\dots,C\}\),
\(M[i,c] \in \mathbb{R}^d\) is the mean representation of item \(i\) in
class \(c\) (Figure~\ref{fig-grid}). The three components are read off
that grid (Figure~\ref{fig-replication}).

\newpage{}

\section{Validating the Measure}\label{sec-appendix-validate}

Experiment 1 defines a ground truth language; however, it cannot
distinguish a failure of the measure from the model representing the
language differently than the structure it was given. Here we remove
that ambiguity by constructing representations in which \(\mu\),
\(\alpha\), \(\beta\), and \(\gamma\) are set directly, with no model or
training, and passing the result through the same analysis used for the
language models. Ground truth therefore exists at the level where the
measure operates.

Specifically, we generated representations with known values of \(\mu\),
\(\alpha\), \(\beta\), and \(\gamma\), varying the number of items,
classes, representation dimensions, observations per word and class, and
noise level across the range used in the experiments. Components were
centered to satisfy their defining constraints and rescaled to
predetermined shares of the total squared length. The grid included the
same factorial Experiment 1 run: the three components measured in
Experiment 2, item counts from 27 to 180, two and four classes, widths
from 8 to 2048, 10, 24, and 100 observations per item and class, three
noise levels, and five seeds. We also generated orthogonal and known
non-orthogonal item and class effects: the orthogonal condition draws
\(\alpha\) and \(\beta\) independently, while the non-orthogonal
condition places a known fraction of \(\beta\) along \(\alpha\)'s
largest dimension. In total, this gave 252,000 runs.

At 24 observations per item and class, the measured interaction matches
the one it was given across the whole range (Table
\ref{tbl-val-recovery}), and is flagged in essentially every run where
anything was present and about one in twenty where nothing was.

\begin{table}

\centering{

\centering\begingroup\fontsize{9}{11}\selectfont

\begin{tabular}[t]{rrrr}
\toprule
specified & runs & measured & flagged\\
\midrule
0.000 & 6300 & 0.0000 [0.0000, 0.0094] & 5\%\\
0.030 & 900 & 0.0323 [0.0199, 0.0418] & 100\%\\
0.048 & 1800 & 0.0495 [0.0385, 0.0688] & 100\%\\
0.056 & 1800 & 0.0573 [0.0458, 0.0918] & 100\%\\
0.059 & 1800 & 0.0607 [0.0495, 0.1041] & 100\%\\
0.100 & 1800 & 0.1007 [0.0911, 0.1089] & 100\%\\
0.111 & 900 & 0.1154 [0.1031, 0.1279] & 100\%\\
0.167 & 1800 & 0.1708 [0.1594, 0.2204] & 100\%\\
0.190 & 1800 & 0.1937 [0.1825, 0.2736] & 100\%\\
0.200 & 1800 & 0.2030 [0.1926, 0.2981] & 100\%\\
0.240 & 900 & 0.2416 [0.2327, 0.2488] & 100\%\\
0.333 & 900 & 0.3396 [0.3263, 0.3610] & 100\%\\
0.444 & 1800 & 0.4488 [0.4381, 0.5059] & 100\%\\
0.485 & 1800 & 0.4877 [0.4787, 0.5645] & 100\%\\
0.500 & 1800 & 0.5025 [0.4941, 0.5873] & 100\%\\
0.667 & 900 & 0.6702 [0.6606, 0.6814] & 100\%\\
0.762 & 1800 & 0.7634 [0.7551, 0.7828] & 100\%\\
0.800 & 1800 & 0.8009 [0.7931, 0.8242] & 100\%\\
0.889 & 900 & 0.8895 [0.8809, 0.8972] & 100\%\\
0.941 & 1800 & 0.9412 [0.9339, 0.9471] & 100\%\\
1.000 & 900 & 0.9999 [0.9936, 1.0000] & 100\%\\
\bottomrule
\end{tabular}
\endgroup{}

}

\caption{\label{tbl-val-recovery}Recovery of the interaction at 24
observations per item and class in 768 or more dimensions, pooled over
noise levels, item counts, and widths. \emph{Specified} is the
interaction share built into the generated representations and
\emph{measured} the median share recovered, with the 2.5th and 97.5th
percentiles across runs in brackets; \emph{flagged} is the share of runs
whose re-split interval excludes zero.}

\end{table}%

All three effects read essentially zero when a component is set to zero,
at every sample size, and each is flagged at approximately the nominal
5\% rate (Table \ref{tbl-val-null}).

\begin{table}

\centering{

\centering\begingroup\fontsize{9}{11}\selectfont

\begin{tabular}[t]{lrrrr}
\toprule
effect & observations & median & 95th pct & flagged\\
\midrule
class effect & 10 & 0.00000 & 0.0081 & 6.1\%\\
class effect & 24 & 0.00000 & 0.0034 & 6.5\%\\
class effect & 100 & 0.00000 & 0.0009 & 6.3\%\\
interaction & 10 & 0.00009 & 0.0644 & 6.2\%\\
interaction & 24 & 0.00004 & 0.0309 & 5.9\%\\
interaction & 100 & 0.00002 & 0.0083 & 4.9\%\\
item effect & 10 & 0.00007 & 0.0568 & 4.1\%\\
item effect & 24 & 0.00003 & 0.0255 & 3.6\%\\
item effect & 100 & 0.00000 & 0.0064 & 5.0\%\\
\bottomrule
\end{tabular}
\endgroup{}

}

\caption{\label{tbl-val-null}What the measure reports when an effect is
set to exactly zero, by observations per item and class, pooled over
every other axis of the validation grid. \emph{Median} and \emph{95th
pct} summarize the measured share of that effect across runs, and
\emph{flagged} is the share of runs whose re-split interval excludes
zero, which should sit near the nominal 5\%.}

\end{table}%

The false-positive rate varies somewhat with representation width and
component, particularly at small widths, but remains below 10\%
throughout the validation grid except for the class effect at the
narrowest width, where it reaches 14\% (Table \ref{tbl-val-fpd}). The
language-model analyses use \(d\geq768\), where the rates are closer to
the nominal level.

\begin{table}

\centering{

\centering\begingroup\fontsize{9}{11}\selectfont

\begin{tabular}[t]{rrrr}
\toprule
width d & item & class & interaction\\
\midrule
8 & 3\% & 14\% & 7\%\\
16 & 7\% & 3\% & 5\%\\
32 & 4\% & 4\% & 10\%\\
128 & 6\% & 4\% & 5\%\\
768 & 2\% & 8\% & 3\%\\
1024 & 3\% & 5\% & 3\%\\
2048 & 0\% & 8\% & 8\%\\
\bottomrule
\end{tabular}
\endgroup{}

}

\caption{\label{tbl-val-fpd}How often a component set to exactly zero is
flagged, by width and by effect.}

\end{table}%

\begin{table}

\centering{

\centering\begingroup\fontsize{9}{11}\selectfont

\begin{tabular}[t]{rrr}
\toprule
width d & measured & analytic chance\\
\midrule
8 & 0.17501 [0.06900, 0.40018] & 0.25000\\
16 & 0.10678 [0.03744, 0.20884] & 0.12500\\
32 & 0.05609 [0.01864, 0.10454] & 0.06250\\
128 & 0.01714 [0.00450, 0.02611] & 0.02344\\
768 & 0.00261 [0.00079, 0.00472] & 0.00391\\
1024 & 0.00217 [0.00065, 0.00330] & 0.00293\\
2048 & 0.00096 [0.00031, 0.00165] & 0.00146\\
\bottomrule
\end{tabular}
\endgroup{}

}

\caption{\label{tbl-val-overlap}Item-class overlap when the two effects
are constructed orthogonal, at 24 observations per item and class.
\emph{Measured} is the median overlap across the runs at that width with
the 2.5th and 97.5th percentiles in brackets; \emph{chance} is the
overlap expected if the component were randomly oriented, \(r/d\) for a
target subspace of \(r\) dimensions in a representation of width \(d\).}

\end{table}%

With orthogonal item and class effects, the measured overlap tracks the
chance level across representation widths (Table \ref{tbl-val-overlap}).
When a known fraction of the class direction is placed inside the item
subspace, the measured overlap rises above chance and is detected in
89\% of runs at \(d=2048\) (Table \ref{tbl-val-power}). Thus, the
measure recovers both the expected chance overlap under orthogonality
and genuine non-orthogonality.

\begin{table}

\centering{

\centering\begingroup\fontsize{9}{11}\selectfont

\begin{tabular}[t]{rrrr}
\toprule
width d & measured & chance & detected\\
\midrule
8 & 0.1688 [0.0824, 0.3998] & 0.12500 & 9\%\\
16 & 0.0951 [0.0414, 0.2026] & 0.06250 & 22\%\\
32 & 0.0593 [0.0225, 0.1119] & 0.03125 & 36\%\\
128 & 0.0213 [0.0062, 0.0517] & 0.00781 & 63\%\\
768 & 0.0065 [0.0011, 0.0298] & 0.00260 & 83\%\\
1024 & 0.0056 [0.0008, 0.0331] & 0.00098 & 82\%\\
2048 & 0.0036 [0.0004, 0.0290] & 0.00049 & 89\%\\
\bottomrule
\end{tabular}
\endgroup{}

}

\caption{\label{tbl-val-power}Measured item-class overlap when a
fraction of the class direction is placed inside the item subspace
(non-orthogonal), at 24 observations per item and class. \emph{Measured}
is the median overlap across the runs at that width with the 2.5th and
97.5th percentiles in brackets; \emph{chance} is the overlap expected if
the component were randomly oriented, \(r/d\) for a target subspace of
\(r\) dimensions in a representation of width \(d\); \emph{detected} is
the fraction of runs whose overlap falls outside that null's 95\%
interval.}

\end{table}%

\subsection{Variance across contexts}\label{sec-appendix-spread}

The interaction is a residual, and a residual absorbs whatever noise the
terms before it do not. A word's average representation in one class is
taken over the contexts it appears in there, so it contains some amount
of sampling error; estimated within a single split, the interaction
would inherit that error and grow with the variance across contexts
whether or not any interaction was present.

The cross-split dot product eliminates this bias. The two estimates come
from disjoint halves of a word's tokens in that class, so their errors
are independent and mean-zero, and the expectation of their inner
product is the true squared length with no variance term: the context
variance enters the variance of the estimate and not its expectation.
This holds for any variance, and for a variance that differs from one
word and class to the next.

Table \ref{tbl-val-noise} demonstrates this. Holding the planted
interaction fixed and raising the variance across contexts to fifty
times the variance between words and classes leaves the measured value
where it is, and where nothing was planted the false-positive rate stays
at the nominal level rather than rising with that variance. If anything
the measure drifts slightly low at the largest variance.

\begin{table}

\centering{

\centering\begingroup\fontsize{8}{10}\selectfont

\begin{tabular}[t]{rrrrr}
\toprule
specified & noise & runs & measured & flagged\\
\midrule
 & 1x & 6300 & 0.0000 & 6\%\\

 & 10x & 6300 & 0.0000 & 6\%\\

\multirow[t]{-3}{*}{\raggedleft\arraybackslash 0.00} & 50x & 6300 & 0.0001 & 6\%\\
\cmidrule{1-5}
 & 1x & 1800 & 0.1012 & 100\%\\

 & 10x & 1800 & 0.1002 & 100\%\\

\multirow[t]{-3}{*}{\raggedleft\arraybackslash 0.10} & 50x & 1800 & 0.1001 & 100\%\\
\cmidrule{1-5}
 & 1x & 1800 & 0.2199 & 100\%\\

 & 10x & 1800 & 0.2049 & 100\%\\

\multirow[t]{-3}{*}{\raggedleft\arraybackslash 0.20} & 50x & 1800 & 0.2016 & 100\%\\
\bottomrule
\end{tabular}
\endgroup{}

}

\caption{\label{tbl-val-noise}Recovery of the interaction across
simulations with different amounts of noise, in 768 or more dimensions,
pooled over item counts, widths and sample sizes. \emph{Specified} is
the interaction share built into the representations and \emph{noise}
the variance across contexts per observation as a multiple of the
variance between words and classes, so at 1x the two are equal.
\emph{Measured} is the median interaction share recovered across runs
and \emph{flagged} the share whose re-split interval excludes zero,
which should sit at 5\% in the rows where nothing was planted.}

\end{table}%

\newpage{}

\section{Width and Orthogonality}\label{sec-appendix-capacity}

\begin{figure}

\centering{

\includegraphics[width=1\linewidth,height=\textheight,keepaspectratio]{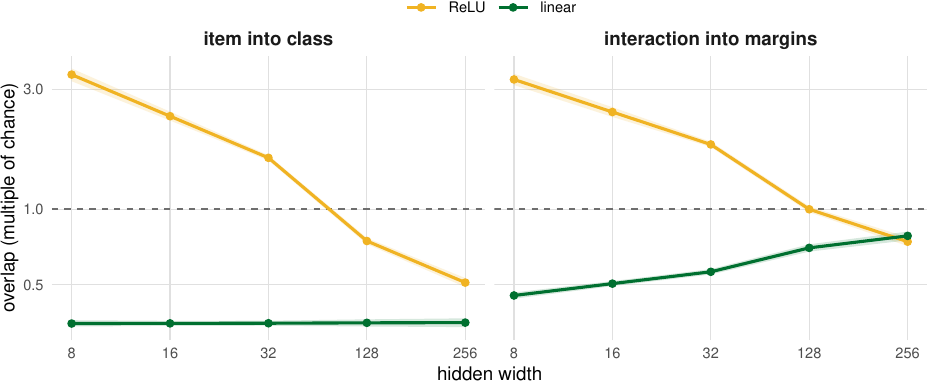}

}

\caption{\label{fig-toy-overlap}Two overlap measurements as a function
of width of the hidden layer. Left: the item effect projected onto the
class direction. Right: the interaction projected onto the span of both
margins. The dashed line is chance.}

\end{figure}

Whether the item and class effects are orthogonal depends on the width
of the representation. Figure~\ref{fig-toy-overlap} reports that
relationship from the fitted models; this appendix gives the remaining
properties of the artificial-language runs.

Under ReLU, overlap stays below chance at wide layers and rises above it
as the layer narrows, while the linear condition stays well below chance
at every width.

The two summaries that follow are descriptive medians across runs rather
than estimates from the fitted models. At low widths, the
representations vary substantially across runs, with this variability
decreasing as width increases. For example, two runs of the same
language differ by 0.70 in the item effect and 0.73 in the class effect
at width 8, compared with 0.16 and 0.15 at width 32. This variability
reflects the different ways models can distribute information when
representational capacity is limited.

Two properties of the language also affect the composition of the
item-by-class structure (Table \ref{tbl-lang}). Increasing the effect of
context reduces the interaction share, while increasing the item's share
in the language produces a corresponding increase in its share of the
representation.

\begin{table}

\centering{

\centering\begingroup\fontsize{8}{10}\selectfont

\begin{tabular}[t]{llrrrr}
\toprule
property & level & runs & item & class & interaction\\
\midrule
 & under 30\% & 3600 & 0.490 & 0.399 & 0.028\\

 & 30 to 70\% & 2871 & 0.538 & 0.371 & 0.008\\

\multirow[t]{-3}{*}{\raggedright\arraybackslash context in the language} & over 70\% & 3886 & 0.578 & 0.350 & 0.004\\
\cmidrule{1-6}
 & item < class & 3525 & 0.428 & 0.480 & 0.017\\

 & item = class & 3427 & 0.569 & 0.359 & 0.015\\

\multirow[t]{-3}{*}{\raggedright\arraybackslash item against class} & item > class & 3405 & 0.644 & 0.294 & 0.011\\
\bottomrule
\end{tabular}
\endgroup{}

}

\caption{\label{tbl-lang}How two other properties of the artificial
language move the composition of the item-by-class structure, under
ReLU, pooled over widths. Rows within a property partition the same
runs. Values are medians of the three shares over the runs in the row.}

\end{table}%

\subsection{Component models}\label{sec-appendix-component-models}

\begin{figure}

\centering{

\includegraphics[width=1\linewidth,height=\textheight,keepaspectratio]{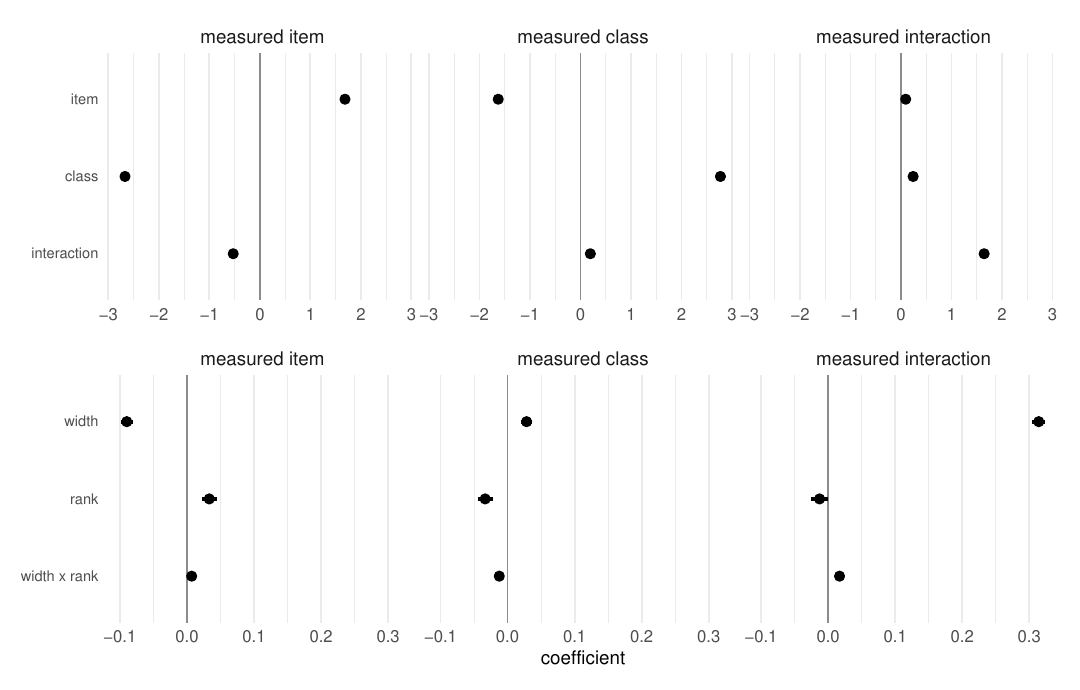}

}

\caption{\label{fig-toy-coefs}Coefficient estimates from the size model
(Equation~\ref{eq-model}), one panel per measured component. The upper
block holds the planted effect sizes and the lower the architecture
terms, which are on a much smaller scale and so are drawn on their own
axis. Points are posterior medians with 95\% intervals.}

\end{figure}

Results for our statistical analysis are presented in Tables
\ref{tbl-component-coefs} and \ref{tbl-component-coefs-lin}, and
visualized in Figure~\ref{fig-toy-coefs}.

\begin{table}

\centering{

\centering\begingroup\fontsize{7}{9}\selectfont

\begin{tabular}[t]{llrrrr}
\toprule
component & term & estimate & error & 95\% CI & P(b > 0)\\
\midrule
item & item & 1.68 & 0.04 & {}[1.61, 1.75] & 1.00\\
\cmidrule{1-6}
class & item & -1.63 & 0.04 & {}[-1.70, -1.56] & 0.00\\
\cmidrule{1-6}
interaction & item & 0.09 & 0.04 & {}[0.02, 0.16] & 0.99\\
\cmidrule{1-6}
item & class & -2.67 & 0.04 & {}[-2.75, -2.59] & 0.00\\
\cmidrule{1-6}
class & class & 2.77 & 0.04 & {}[2.69, 2.85] & 1.00\\
\cmidrule{1-6}
interaction & class & 0.24 & 0.04 & {}[0.16, 0.31] & 1.00\\
\cmidrule{1-6}
item & interaction & -0.53 & 0.03 & {}[-0.60, -0.46] & 0.00\\
\cmidrule{1-6}
class & interaction & 0.20 & 0.03 & {}[0.13, 0.26] & 1.00\\
\cmidrule{1-6}
interaction & interaction & 1.65 & 0.03 & {}[1.58, 1.71] & 1.00\\
\cmidrule{1-6}
item & width & -0.09 & 0.00 & {}[-0.10, -0.08] & 0.00\\
\cmidrule{1-6}
class & width & 0.03 & 0.00 & {}[0.02, 0.04] & 1.00\\
\cmidrule{1-6}
interaction & width & 0.31 & 0.00 & {}[0.30, 0.32] & 1.00\\
\cmidrule{1-6}
item & rank & 0.03 & 0.01 & {}[0.02, 0.04] & 1.00\\
\cmidrule{1-6}
class & rank & -0.03 & 0.01 & {}[-0.04, -0.02] & 0.00\\
\cmidrule{1-6}
interaction & rank & -0.01 & 0.01 & {}[-0.03, -0.00] & 0.02\\
\cmidrule{1-6}
item & width x rank & 0.01 & 0.00 & {}[0.00, 0.01] & 0.99\\
\cmidrule{1-6}
class & width x rank & -0.01 & 0.00 & {}[-0.02, -0.01] & 0.00\\
\cmidrule{1-6}
interaction & width x rank & 0.02 & 0.00 & {}[0.01, 0.02] & 1.00\\
\cmidrule{1-6}
item & item x width & -0.00 & 0.02 & {}[-0.04, 0.04] & 0.47\\
\cmidrule{1-6}
class & item x width & 0.01 & 0.02 & {}[-0.02, 0.05] & 0.76\\
\cmidrule{1-6}
interaction & item x width & -0.14 & 0.02 & {}[-0.18, -0.10] & 0.00\\
\cmidrule{1-6}
item & class x width & 0.48 & 0.02 & {}[0.44, 0.52] & 1.00\\
\cmidrule{1-6}
class & class x width & -0.54 & 0.02 & {}[-0.58, -0.50] & 0.00\\
\cmidrule{1-6}
interaction & class x width & -0.12 & 0.02 & {}[-0.16, -0.07] & 0.00\\
\cmidrule{1-6}
item & interaction x width & -0.20 & 0.02 & {}[-0.24, -0.16] & 0.00\\
\cmidrule{1-6}
class & interaction x width & 0.09 & 0.02 & {}[0.05, 0.13] & 1.00\\
\cmidrule{1-6}
interaction & interaction x width & -0.11 & 0.02 & {}[-0.15, -0.07] & 0.00\\
\cmidrule{1-6}
item & item x rank & -0.17 & 0.03 & {}[-0.22, -0.12] & 0.00\\
\cmidrule{1-6}
class & item x rank & 0.16 & 0.03 & {}[0.11, 0.21] & 1.00\\
\cmidrule{1-6}
interaction & item x rank & 0.09 & 0.03 & {}[0.03, 0.14] & 1.00\\
\cmidrule{1-6}
item & class x rank & 0.04 & 0.03 & {}[-0.02, 0.10] & 0.90\\
\cmidrule{1-6}
class & class x rank & -0.06 & 0.03 & {}[-0.12, -0.01] & 0.02\\
\cmidrule{1-6}
interaction & class x rank & 0.04 & 0.03 & {}[-0.02, 0.10] & 0.91\\
\cmidrule{1-6}
item & interaction x rank & 0.01 & 0.03 & {}[-0.04, 0.06] & 0.64\\
\cmidrule{1-6}
class & interaction x rank & -0.04 & 0.03 & {}[-0.09, 0.00] & 0.04\\
\cmidrule{1-6}
interaction & interaction x rank & 0.06 & 0.03 & {}[0.02, 0.11] & 0.99\\
\cmidrule{1-6}
item & item x width x rank & -0.05 & 0.01 & {}[-0.08, -0.02] & 0.00\\
\cmidrule{1-6}
class & item x width x rank & 0.05 & 0.01 & {}[0.03, 0.08] & 1.00\\
\cmidrule{1-6}
interaction & item x width x rank & -0.02 & 0.01 & {}[-0.05, 0.00] & 0.05\\
\cmidrule{1-6}
item & class x width x rank & 0.01 & 0.02 & {}[-0.02, 0.04] & 0.70\\
\cmidrule{1-6}
class & class x width x rank & -0.01 & 0.02 & {}[-0.04, 0.02] & 0.29\\
\cmidrule{1-6}
interaction & class x width x rank & -0.02 & 0.02 & {}[-0.05, 0.01] & 0.12\\
\cmidrule{1-6}
item & interaction x width x rank & -0.03 & 0.01 & {}[-0.05, -0.00] & 0.02\\
\cmidrule{1-6}
class & interaction x width x rank & 0.03 & 0.01 & {}[0.01, 0.06] & 0.99\\
\cmidrule{1-6}
interaction & interaction x width x rank & -0.02 & 0.01 & {}[-0.05, 0.01] & 0.09\\
\bottomrule
\end{tabular}
\endgroup{}

}

\caption{\label{tbl-component-coefs}Coefficient values of the size model
(Equation~\ref{eq-model}) with error and 95\% CIs, ReLU arm.}

\end{table}%

\begin{table}

\centering{

\centering\begingroup\fontsize{7}{9}\selectfont

\begin{tabular}[t]{llrrrr}
\toprule
component & term & estimate & error & 95\% CI & P(b > 0)\\
\midrule
item & item & 2.36 & 0.04 & {}[2.29, 2.43] & 1.00\\
\cmidrule{1-6}
class & item & -2.29 & 0.03 & {}[-2.35, -2.22] & 0.00\\
\cmidrule{1-6}
interaction & item & 0.48 & 0.04 & {}[0.40, 0.55] & 1.00\\
\cmidrule{1-6}
item & class & -3.52 & 0.04 & {}[-3.60, -3.45] & 0.00\\
\cmidrule{1-6}
class & class & 3.66 & 0.04 & {}[3.58, 3.74] & 1.00\\
\cmidrule{1-6}
interaction & class & 0.08 & 0.04 & {}[-0.00, 0.16] & 0.97\\
\cmidrule{1-6}
item & interaction & 0.94 & 0.03 & {}[0.88, 1.00] & 1.00\\
\cmidrule{1-6}
class & interaction & -1.05 & 0.03 & {}[-1.11, -0.99] & 0.00\\
\cmidrule{1-6}
interaction & interaction & 1.57 & 0.03 & {}[1.50, 1.64] & 1.00\\
\cmidrule{1-6}
item & width & 0.12 & 0.00 & {}[0.12, 0.13] & 1.00\\
\cmidrule{1-6}
class & width & -0.12 & 0.00 & {}[-0.13, -0.12] & 0.00\\
\cmidrule{1-6}
interaction & width & 0.16 & 0.01 & {}[0.15, 0.17] & 1.00\\
\cmidrule{1-6}
item & rank & 0.02 & 0.01 & {}[0.01, 0.03] & 1.00\\
\cmidrule{1-6}
class & rank & -0.01 & 0.01 & {}[-0.02, 0.00] & 0.04\\
\cmidrule{1-6}
interaction & rank & -0.08 & 0.01 & {}[-0.09, -0.06] & 0.00\\
\cmidrule{1-6}
item & width x rank & 0.00 & 0.00 & {}[-0.00, 0.01] & 0.87\\
\cmidrule{1-6}
class & width x rank & -0.01 & 0.00 & {}[-0.01, -0.00] & 0.01\\
\cmidrule{1-6}
interaction & width x rank & 0.02 & 0.00 & {}[0.02, 0.03] & 1.00\\
\cmidrule{1-6}
item & item x width & -0.45 & 0.02 & {}[-0.49, -0.42] & 0.00\\
\cmidrule{1-6}
class & item x width & 0.38 & 0.02 & {}[0.35, 0.42] & 1.00\\
\cmidrule{1-6}
interaction & item x width & -0.02 & 0.02 & {}[-0.06, 0.02] & 0.11\\
\cmidrule{1-6}
item & class x width & 0.36 & 0.02 & {}[0.32, 0.40] & 1.00\\
\cmidrule{1-6}
class & class x width & -0.43 & 0.02 & {}[-0.47, -0.39] & 0.00\\
\cmidrule{1-6}
interaction & class x width & -0.08 & 0.02 & {}[-0.13, -0.04] & 0.00\\
\cmidrule{1-6}
item & interaction x width & -0.13 & 0.02 & {}[-0.16, -0.09] & 0.00\\
\cmidrule{1-6}
class & interaction x width & 0.04 & 0.02 & {}[0.00, 0.07] & 0.99\\
\cmidrule{1-6}
interaction & interaction x width & 0.10 & 0.02 & {}[0.07, 0.14] & 1.00\\
\cmidrule{1-6}
item & item x rank & -0.04 & 0.03 & {}[-0.09, 0.01] & 0.05\\
\cmidrule{1-6}
class & item x rank & 0.03 & 0.03 & {}[-0.02, 0.08] & 0.90\\
\cmidrule{1-6}
interaction & item x rank & 0.19 & 0.03 & {}[0.14, 0.24] & 1.00\\
\cmidrule{1-6}
item & class x rank & -0.04 & 0.03 & {}[-0.10, 0.01] & 0.07\\
\cmidrule{1-6}
class & class x rank & 0.00 & 0.03 & {}[-0.05, 0.06] & 0.56\\
\cmidrule{1-6}
interaction & class x rank & 0.13 & 0.03 & {}[0.07, 0.19] & 1.00\\
\cmidrule{1-6}
item & interaction x rank & -0.11 & 0.02 & {}[-0.15, -0.06] & 0.00\\
\cmidrule{1-6}
class & interaction x rank & 0.07 & 0.02 & {}[0.02, 0.11] & 1.00\\
\cmidrule{1-6}
interaction & interaction x rank & 0.18 & 0.02 & {}[0.13, 0.23] & 1.00\\
\cmidrule{1-6}
item & item x width x rank & -0.00 & 0.01 & {}[-0.03, 0.02] & 0.36\\
\cmidrule{1-6}
class & item x width x rank & 0.01 & 0.01 & {}[-0.01, 0.04] & 0.85\\
\cmidrule{1-6}
interaction & item x width x rank & -0.05 & 0.01 & {}[-0.08, -0.02] & 0.00\\
\cmidrule{1-6}
item & class x width x rank & 0.02 & 0.01 & {}[-0.01, 0.05] & 0.92\\
\cmidrule{1-6}
class & class x width x rank & -0.01 & 0.01 & {}[-0.04, 0.02] & 0.26\\
\cmidrule{1-6}
interaction & class x width x rank & -0.05 & 0.02 & {}[-0.08, -0.02] & 0.00\\
\cmidrule{1-6}
item & interaction x width x rank & 0.00 & 0.01 & {}[-0.02, 0.03] & 0.58\\
\cmidrule{1-6}
class & interaction x width x rank & -0.00 & 0.01 & {}[-0.03, 0.02] & 0.40\\
\cmidrule{1-6}
interaction & interaction x width x rank & -0.00 & 0.01 & {}[-0.03, 0.02] & 0.36\\
\bottomrule
\end{tabular}
\endgroup{}

}

\caption{\label{tbl-component-coefs-lin}Coefficient values of the size
model (Equation~\ref{eq-model}) with error and 95\% CIs, linear arm.}

\end{table}%

\subsection{Overlap models}\label{sec-appendix-overlap-models}

Table \ref{tbl-overlap-coefs} reports every coefficient of the overlap
model (Equation~\ref{eq-model}).

\begin{table}

\centering{

\centering\begingroup\fontsize{7}{9}\selectfont

\begin{tabular}[t]{lllrrrr}
\toprule
arm & measure & term & estimate & error & 95\% CI & P(b > 0)\\
\midrule
 & item\_class & item & -0.51 & 0.04 & {}[-0.59, -0.42] & 0.00\\

 & int\_margins & item & -0.96 & 0.04 & {}[-1.05, -0.88] & 0.00\\

 & item\_class & class & -1.62 & 0.05 & {}[-1.71, -1.53] & 0.00\\

 & int\_margins & class & -0.62 & 0.05 & {}[-0.71, -0.53] & 0.00\\

 & item\_class & interaction & 0.31 & 0.04 & {}[0.23, 0.39] & 1.00\\

 & int\_margins & interaction & -0.57 & 0.04 & {}[-0.65, -0.49] & 0.00\\

 & item\_class & width & -0.38 & 0.01 & {}[-0.40, -0.37] & 0.00\\

 & int\_margins & width & -0.30 & 0.01 & {}[-0.32, -0.29] & 0.00\\

 & item\_class & rank & -0.03 & 0.01 & {}[-0.04, -0.01] & 0.00\\

 & int\_margins & rank & -0.03 & 0.01 & {}[-0.05, -0.02] & 0.00\\

 & item\_class & width x rank & -0.01 & 0.01 & {}[-0.03, -0.00] & 0.00\\

 & int\_margins & width x rank & -0.02 & 0.01 & {}[-0.04, -0.01] & 0.00\\

 & item\_class & item x width & -0.30 & 0.03 & {}[-0.37, -0.23] & 0.00\\

 & int\_margins & item x width & -0.39 & 0.03 & {}[-0.46, -0.33] & 0.00\\

 & item\_class & class x width & 0.37 & 0.04 & {}[0.29, 0.44] & 1.00\\

 & int\_margins & class x width & 0.12 & 0.04 & {}[0.04, 0.19] & 1.00\\

 & item\_class & interaction x width & 0.34 & 0.03 & {}[0.27, 0.40] & 1.00\\

 & int\_margins & interaction x width & 0.04 & 0.03 & {}[-0.03, 0.10] & 0.85\\

 & item\_class & item x rank & 0.12 & 0.03 & {}[0.06, 0.18] & 1.00\\

 & int\_margins & item x rank & 0.09 & 0.03 & {}[0.03, 0.15] & 1.00\\

 & item\_class & class x rank & 0.12 & 0.03 & {}[0.05, 0.18] & 1.00\\

 & int\_margins & class x rank & 0.04 & 0.03 & {}[-0.03, 0.11] & 0.88\\

 & item\_class & interaction x rank & 0.09 & 0.03 & {}[0.03, 0.15] & 1.00\\

 & int\_margins & interaction x rank & 0.04 & 0.03 & {}[-0.02, 0.10] & 0.92\\

 & item\_class & item x width x rank & 0.05 & 0.02 & {}[0.00, 0.10] & 0.98\\

 & int\_margins & item x width x rank & 0.06 & 0.02 & {}[0.01, 0.11] & 0.99\\

 & item\_class & class x width x rank & 0.02 & 0.03 & {}[-0.03, 0.07] & 0.77\\

 & int\_margins & class x width x rank & 0.01 & 0.03 & {}[-0.04, 0.07] & 0.69\\

 & item\_class & interaction x width x rank & -0.03 & 0.02 & {}[-0.08, 0.02] & 0.13\\

\multirow[t]{-30}{*}{\raggedright\arraybackslash relu} & int\_margins & interaction x width x rank & -0.04 & 0.03 & {}[-0.09, 0.01] & 0.04\\
\cmidrule{1-7}
 & item\_class & item & -0.57 & 0.05 & {}[-0.67, -0.47] & 0.00\\

 & int\_margins & item & -1.71 & 0.05 & {}[-1.81, -1.61] & 0.00\\

 & item\_class & class & -0.49 & 0.05 & {}[-0.59, -0.38] & 0.00\\

 & int\_margins & class & -0.69 & 0.05 & {}[-0.80, -0.58] & 0.00\\

 & item\_class & interaction & 0.12 & 0.05 & {}[0.02, 0.22] & 0.99\\

 & int\_margins & interaction & -0.42 & 0.05 & {}[-0.52, -0.32] & 0.00\\

 & item\_class & width & -0.00 & 0.00 & {}[-0.01, 0.01] & 0.40\\

 & int\_margins & width & 0.11 & 0.00 & {}[0.10, 0.11] & 1.00\\

 & item\_class & rank & -0.05 & 0.01 & {}[-0.06, -0.03] & 0.00\\

 & int\_margins & rank & -0.05 & 0.01 & {}[-0.07, -0.03] & 0.00\\

 & item\_class & width x rank & -0.02 & 0.00 & {}[-0.02, -0.01] & 0.00\\

 & int\_margins & width x rank & -0.02 & 0.00 & {}[-0.03, -0.02] & 0.00\\

 & item\_class & item x width & 0.25 & 0.02 & {}[0.21, 0.29] & 1.00\\

 & int\_margins & item x width & -0.40 & 0.02 & {}[-0.43, -0.36] & 0.00\\

 & item\_class & class x width & 0.31 & 0.02 & {}[0.27, 0.36] & 1.00\\

 & int\_margins & class x width & -0.30 & 0.02 & {}[-0.34, -0.26] & 0.00\\

 & item\_class & interaction x width & -0.10 & 0.02 & {}[-0.13, -0.06] & 0.00\\

 & int\_margins & interaction x width & -0.65 & 0.02 & {}[-0.68, -0.61] & 0.00\\

 & item\_class & item x rank & 0.04 & 0.04 & {}[-0.03, 0.12] & 0.88\\

 & int\_margins & item x rank & 0.07 & 0.04 & {}[-0.01, 0.14] & 0.96\\

 & item\_class & class x rank & 0.06 & 0.04 & {}[-0.01, 0.14] & 0.94\\

 & int\_margins & class x rank & -0.01 & 0.04 & {}[-0.09, 0.07] & 0.40\\

 & item\_class & interaction x rank & -0.23 & 0.04 & {}[-0.30, -0.16] & 0.00\\

 & int\_margins & interaction x rank & -0.26 & 0.04 & {}[-0.33, -0.19] & 0.00\\

 & item\_class & item x width x rank & 0.01 & 0.01 & {}[-0.02, 0.03] & 0.67\\

 & int\_margins & item x width x rank & 0.01 & 0.01 & {}[-0.01, 0.04] & 0.84\\

 & item\_class & class x width x rank & 0.00 & 0.02 & {}[-0.03, 0.03] & 0.51\\

 & int\_margins & class x width x rank & -0.02 & 0.02 & {}[-0.05, 0.01] & 0.07\\

 & item\_class & interaction x width x rank & -0.09 & 0.01 & {}[-0.12, -0.07] & 0.00\\

\multirow[t]{-30}{*}{\raggedright\arraybackslash linear} & int\_margins & interaction x width x rank & -0.17 & 0.01 & {}[-0.19, -0.14] & 0.00\\
\bottomrule
\end{tabular}
\endgroup{}

}

\caption{\label{tbl-overlap-coefs}Every coefficient of the overlap model
(Equation~\ref{eq-model}), with Estimate, Estimated error, and 95\%
CIs.}

\end{table}%

\begin{figure}

\centering{

\includegraphics[width=1\linewidth,height=\textheight,keepaspectratio]{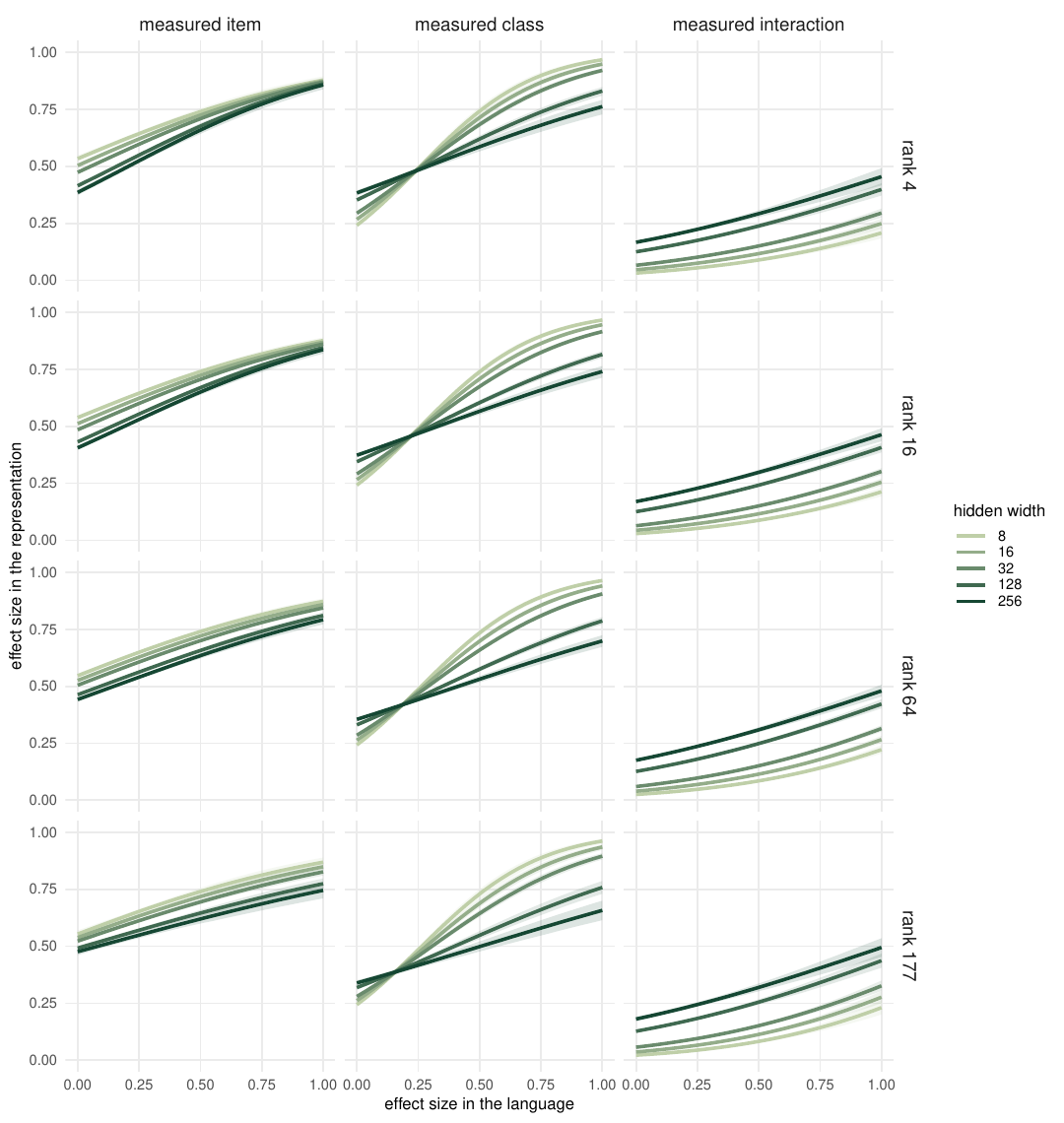}

}

\caption{\label{fig-toy-effects-full}Model results for the size model
(Equation~\ref{eq-model}). Columns are the three measured components,
rows are the planted interaction rank, and lines are hidden widths, with
95\% intervals. Holding rank fixed within a row is what separates the
width effect from the width-by-rank term; the near-identical rows show
how little the rank changes it.}

\end{figure}%

\begin{figure}

\centering{

\includegraphics[width=1\linewidth,height=\textheight,keepaspectratio]{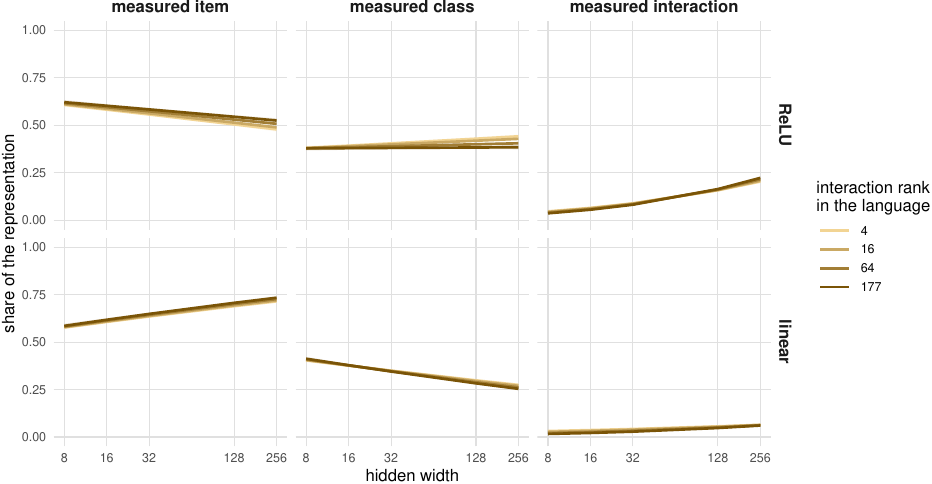}

}

\caption{\label{fig-toy-effects}Fraction of the representation by each
component (such that they sum to 1), against hidden width, with one line
per rank of the interaction planted in the language and 95\% intervals.}

\end{figure}%

\newpage{}

\section{Stimulus Counts}\label{sec-appendix-stimuli}

The six general-domain English Universal Dependencies treebanks
\citep{nivre2020ud} we concatenate are EWT \citep{silveira2014ewt}, GUM
\citep{zeldes2017gum}, LinES \citep{ahrenberg2007lines}, ParTUT
\citep{sanguinetti2015partut}, GENTLE \citep{aoyama2023gentle}, and PUD
\citep{zeman2017pud}.

\begin{table}

\centering{

\centering\begingroup\fontsize{8}{10}\selectfont

\begin{tabular}[t]{lrl}
\toprule
class & item types & median tokens\\
\midrule
part of speech (noun / verb) & 78 & 18 / 18\\
grammatical role (subject / object) & 49 & 15 / 19\\
\bottomrule
\end{tabular}
\endgroup{}

}

\caption{\label{tbl-constructions}Stimuli for the two linguistic
categories, with the two classes of each in parentheses. \emph{Item
types} is how many word forms occur in both of them at least ten times;
forms occurring in only one are excluded. \emph{Median tokens} is how
often a typical form occurs in each of its two classes, in the order
they are listed.}

\end{table}%

\newpage{}

\section{Model Details}\label{sec-appendix-models}

\begin{table}

\centering{

\centering\begingroup\fontsize{9}{11}\selectfont

\begin{tabular}[t]{llrrrlr}
\toprule
family & model & params & width & layers & training data & tokens\\
\midrule
 & opt-babylm-125m & 125M & 768 & 12 & BabyLM (child-available) & \textasciitilde{}3B\\

 & opt-babylm-350m & 350M & 1024 & 24 & BabyLM (child-available) & \textasciitilde{}3B\\

\multirow{-3}{*}{\raggedright\arraybackslash BabyLM} & opt-babylm-1.3B & 1.3B & 2048 & 24 & BabyLM (child-available) & \textasciitilde{}3B\\
\cmidrule{1-7}
 & pythia-160m & 160M & 768 & 12 & the Pile & \textasciitilde{}300B\\

 & pythia-410m & 410M & 1024 & 24 & the Pile & \textasciitilde{}300B\\

\multirow{-3}{*}{\raggedright\arraybackslash Pythia} & pythia-1.4b & 1.4B & 2048 & 24 & the Pile & \textasciitilde{}300B\\
\bottomrule
\end{tabular}
\endgroup{}

}

\caption{\label{tbl-models}The six models, paired by width.
Training-token counts are approximate; the OPT-BabyLM models train for
20 epochs over a 100M-word corpus, which is roughly 150M tokens, so
their effective token exposure is about twenty times that.}

\end{table}%

The six models are listed in Table~\ref{tbl-models}. The two families
are matched on width rather than on parameter count, which is what the
measure is sensitive to: the chance level for overlap is the class
subspace rank over the width, so a comparison across models of different
width has to be read against different chance levels. Pairing at 768,
1024, and 2048 removes that from the family comparison.

Beyond width, the families differ in the size of their training corpora,
by more than two orders of magnitude, which is the manipulation of
interest.

\newpage{}

\section{Standardizing the Hidden
States}\label{sec-appendix-standardize}

Rogue dimensions could in principle affect what we measure. Table
\ref{tbl-standardize} reports part of speech at the deepest layer,
measured both with and without the per-dimension rescaling.

\begin{table}

\centering{

\centering\begingroup\fontsize{7}{9}\selectfont

\begin{tabular}[t]{lrrrrrr}
\toprule
\multicolumn{1}{c}{ } & \multicolumn{2}{c}{interaction} & \multicolumn{2}{c}{class effect} & \multicolumn{2}{c}{item-class overlap} \\
\cmidrule(l{3pt}r{3pt}){2-3} \cmidrule(l{3pt}r{3pt}){4-5} \cmidrule(l{3pt}r{3pt}){6-7}
model & std & raw & std & raw & std & raw\\
\midrule
BabyLM 125M & 0.095 & 0.095 & 0.072 & 0.074 & 0.0052 & 0.0053\\
BabyLM 350M & 0.122 & 0.122 & 0.066 & 0.068 & 0.0056 & 0.0061\\
BabyLM 1.3B & 0.104 & 0.104 & 0.050 & 0.051 & 0.0035 & 0.0036\\
Pythia 160M & 0.231 & 0.228 & 0.078 & 0.089 & 0.0058 & 0.0081\\
Pythia 410M & 0.244 & 0.242 & 0.098 & 0.123 & 0.0067 & 0.0195\\
Pythia 1.4B & 0.242 & 0.238 & 0.086 & 0.124 & 0.0055 & 0.0234\\
\bottomrule
\end{tabular}
\endgroup{}

}

\caption{\label{tbl-standardize}Part of speech at the deepest layer,
measured with and without the per-dimension rescaling. Sizes are shares
of the item-by-class structure; overlaps are the raw fractions, not
multiples of chance.}

\end{table}%

\newpage{}

\section{Deepest-Layer Measurements}\label{sec-appendix-llm}

Table \ref{tbl-llm} gives the results of the overlap measurement at the
final layer along with the null.

\begin{table}

\centering{

\centering\begingroup\fontsize{7}{9}\selectfont

\begin{tabular}[t]{llrrrr}
\toprule
class & model & item-class overlap & its null & interaction overlap & its null\\
\midrule
 & BabyLM 125M & 0.0052 (4.2×) & 0.0012 & 0.178 (3.5×) & 0.051\\

 & BabyLM 350M & 0.0056 (5.9×) & 0.0010 & 0.166 (4.3×) & 0.039\\

 & BabyLM 1.3B & 0.0035 (7.6×) & 0.0005 & 0.106 (4.8×) & 0.022\\

 & Pythia 160M & 0.0058 (4.6×) & 0.0013 & 0.231 (4.2×) & 0.055\\

 & Pythia 410M & 0.0067 (6.9×) & 0.0010 & 0.239 (6.1×) & 0.039\\

\multirow[t]{-6}{*}{\raggedright\arraybackslash part of speech} & Pythia 1.4B & 0.0055 (11.3×) & 0.0005 & 0.209 (9.7×) & 0.021\\
\cmidrule{1-6}
 & BabyLM 125M & 0.0054 (4.0×) & 0.0013 & 0.118 (3.5×) & 0.034\\

 & BabyLM 350M & 0.0051 (5.5×) & 0.0009 & 0.102 (4.0×) & 0.025\\

 & BabyLM 1.3B & 0.0033 (6.7×) & 0.0005 & 0.071 (4.7×) & 0.015\\

 & Pythia 160M & 0.0049 (3.9×) & 0.0012 & 0.179 (5.3×) & 0.034\\

 & Pythia 410M & 0.0047 (4.9×) & 0.0010 & 0.161 (6.5×) & 0.025\\

\multirow[t]{-6}{*}{\raggedright\arraybackslash grammatical role} & Pythia 1.4B & 0.0062 (13.0×) & 0.0005 & 0.141 (11.2×) & 0.013\\
\bottomrule
\end{tabular}
\endgroup{}

}

\caption{\label{tbl-llm}Overlap measurement and the null values at the
final layer of each model. Each overlap is the fraction of one component
lying inside another's subspace: the item effect inside the class
subspace, the interaction inside the span of the two main effects. Each
is followed by the overlap divided by its own null, so 4.2x means the
overlap is 4.2 times what random orientation gives.}

\end{table}%

Table \ref{tbl-output} reports each component's share at the final
hidden layer alongside the same measurement after the unembedding, on
the same tokens. A hidden layer is only meaningful relative to what
reads it, and a linear map whose gain differs across directions does not
preserve shares, so the two need not agree. The interaction's share is
close to stable in both categories, moving by at most 0.045, so the
claims we make about it do not depend on where it was measured. The main
effects shift somewhat for part of speech, where the class effect takes
a larger share of the structure after the unembedding than it holds in
the hidden layer.

\begin{table}

\centering{

\centering\begingroup\fontsize{7}{9}\selectfont

\begin{tabular}[t]{llrrrrrr}
\toprule
\multicolumn{2}{c}{ } & \multicolumn{2}{c}{item effect} & \multicolumn{2}{c}{class effect} & \multicolumn{2}{c}{interaction} \\
\cmidrule(l{3pt}r{3pt}){3-4} \cmidrule(l{3pt}r{3pt}){5-6} \cmidrule(l{3pt}r{3pt}){7-8}
class & model & hidden & output & hidden & output & hidden & output\\
\midrule
 & BabyLM 125M & 0.832 & 0.751 & 0.069 & 0.142 & 0.099 & 0.107\\

 & BabyLM 350M & 0.810 & 0.721 & 0.064 & 0.146 & 0.126 & 0.134\\

 & BabyLM 1.3B & 0.843 & 0.757 & 0.049 & 0.123 & 0.108 & 0.120\\

 & Pythia 160M & 0.688 & 0.577 & 0.077 & 0.143 & 0.235 & 0.280\\

 & Pythia 410M & 0.656 & 0.619 & 0.095 & 0.121 & 0.248 & 0.260\\

\multirow[t]{-6}{*}{\raggedright\arraybackslash part of speech} & Pythia 1.4B & 0.670 & 0.614 & 0.084 & 0.135 & 0.246 & 0.251\\
\cmidrule{1-8}
 & BabyLM 125M & 0.903 & 0.901 & 0.050 & 0.057 & 0.047 & 0.042\\

 & BabyLM 350M & 0.890 & 0.887 & 0.050 & 0.062 & 0.060 & 0.051\\

 & BabyLM 1.3B & 0.894 & 0.884 & 0.039 & 0.052 & 0.067 & 0.063\\

 & Pythia 160M & 0.828 & 0.904 & 0.078 & 0.026 & 0.094 & 0.070\\

 & Pythia 410M & 0.838 & 0.844 & 0.074 & 0.071 & 0.088 & 0.084\\

\multirow[t]{-6}{*}{\raggedright\arraybackslash grammatical role} & Pythia 1.4B & 0.828 & 0.830 & 0.078 & 0.081 & 0.094 & 0.089\\
\bottomrule
\end{tabular}
\endgroup{}

}

\caption{\label{tbl-output}Each component's share of the item-by-class
structure at the final hidden layer and after the unembedding, measured
on identical tokens. Both members of a pair are point estimates from one
run, which is what makes them comparable to each other; they therefore
sit about 0.003 from the re-split medians quoted elsewhere.}

\end{table}%

\subsection{How much of a token the components account
for}\label{sec-appendix-gain}

The fraction of squared length reported for each measurement
demonstrates how much a representation comprises item, class, or
interaction components relative to each other. However, it does not say
how much of the entire representation comprises that information. In
this section, we calculate how much of the overall representation each
component takes up (Table \ref{tbl-gain}). We operationalize this by
estimating how much of a token's hidden state each one accounts for as
the reduction in squared error against guessing the grand mean. The
components are orthogonal, so their contributions accumulate, and the
three do not sum to the whole, since most of what varies in a token is
the context it appears in.

\begin{table}

\centering{

\centering\begingroup\fontsize{7}{9}\selectfont

\begin{tabular}[t]{llrrr}
\toprule
class & model & word & class effect & interaction\\
\midrule
 & BabyLM 125M & 36.3 & 3.1 & 4.2\\

 & BabyLM 350M & 31.2 & 2.5 & 4.7\\

 & BabyLM 1.3B & 25.8 & 1.5 & 3.2\\

 & Pythia 160M & 22.4 & 2.5 & 7.5\\

 & Pythia 410M & 20.3 & 3.0 & 7.6\\

\multirow[t]{-6}{*}{\raggedright\arraybackslash part of speech} & Pythia 1.4B & 18.8 & 2.4 & 6.8\\
\cmidrule{1-5}
 & BabyLM 125M & 38.4 & 2.1 & 1.9\\

 & BabyLM 350M & 33.0 & 1.8 & 2.2\\

 & BabyLM 1.3B & 25.9 & 1.1 & 1.9\\

 & Pythia 160M & 28.0 & 2.6 & 3.1\\

 & Pythia 410M & 27.5 & 2.4 & 2.9\\

\multirow[t]{-6}{*}{\raggedright\arraybackslash grammatical role} & Pythia 1.4B & 24.9 & 2.4 & 2.8\\
\bottomrule
\end{tabular}
\endgroup{}

}

\caption{\label{tbl-gain}How much of a token's hidden state each
component accounts for, at the deepest layer, as the percentage
reduction in squared error against guessing the grand mean.}

\end{table}%

\newpage{}

\section{Ruling Out Context}\label{sec-appendix-context}

Because our stimuli use the same tokens across classes, the two uses of
an item at the word position differ only in their left contexts. The
item-by-class interaction could therefore reflect differences in context
rather than lexical structure built by the model. To test this
possibility, we estimate how much of the interaction can be predicted
from the left context alone.

In a causal model, the hidden state at a word position depends on the
current word and the preceding context. We approximate the contribution
of the preceding context using a linear map, \(h_t \approx W h_{t-1}\),
which predicts the next state from the previous one. We fit a separate
map at each layer using empirical-Bayes ridge regression, with the
regularization strength selected by evidence optimization. Crucially, we
train each map on token pairs whose target words are not among the items
in our analysis. The map therefore estimates context carry-over without
being trained on the items to which we apply it.

We then apply this map to the preceding-word states used to compute the
interaction, yielding the interaction predicted by context alone. To
measure the similarity between the actual and predicted interactions, we
calculate their squared cosine similarity. We estimate the overlap
between the two interactions by averaging the dot products between the
predicted interaction from one split and the actual interaction from the
other. We then divide the squared overlap by the product of the two
interactions' sizes, with each size estimated by the dot product between
its two independent splits. Using cross-split dot products ensures that
the estimates reflect shared structure rather than noise specific to
either split. We call the resulting value the \emph{context-explained
share}; its complement, the \emph{word-built share}, is the portion not
explained by this item-independent prediction.

Across all six models and both categories, the context explains only a
small part of the interaction at the deepest layer
(Table~\ref{tbl-residual}). The word-built share runs 0.79--0.92, while
the context-explained share runs 0.08--0.21. A null control that
mismatches items before the comparison yields values indistinguishable
from zero (at most 0.000 in absolute value). At shallower layers, the
context-explained share is larger in the OPT-BabyLM models, reaching
0.44 at a quarter of model depth, but decreases with depth. In Pythia,
it remains small throughout. Thus, the interaction cannot be explained
as a linear prediction from the preceding context alone, using the same
standard of linear decodability applied to the probe in Experiment
3.\footnote{Replacing the linear carry-over map with a nonlinear
  alternative, a one-hidden-layer MLP initialized from the linear map
  and trained on the same token pairs, does not increase the
  context-explained share. The MLP reduces held-out prediction error on
  word states by 5--25\% relative to the linear map (median: about
  13\%), yet changes the context-explained share by no more than 0.03 at
  any layer. The result is therefore unlikely to arise simply because
  the linear map fails to capture context structure.}

A simpler magnitude control yields a consistent result. Applying the
same decomposition to the pre-word state, which represents the context
before the target word is encountered, produces a smaller interaction
relative to the overall representation. This proportion also remains
flat across depth, unlike the interaction at the word position, which
grows with depth.

\begin{table}

\centering{

\centering\begingroup\fontsize{7}{9}\selectfont

\begin{tabular}[t]{lrrr}
\toprule
model & word-built & context & null\\
\midrule
\addlinespace[0.3em]
\multicolumn{4}{l}{\textbf{POS (noun/verb)}}\\
\hspace{1em}BabyLM 125M & 0.85 [0.82, 0.88] & 0.15 & -0.00\\
\hspace{1em}Pythia 160M & 0.86 [0.83, 0.87] & 0.14 & -0.00\\
\hspace{1em}BabyLM 350M & 0.89 [0.86, 0.91] & 0.11 & -0.00\\
\hspace{1em}Pythia 410M & 0.81 [0.77, 0.83] & 0.19 & -0.00\\
\hspace{1em}BabyLM 1.3B & 0.89 [0.87, 0.91] & 0.11 & -0.00\\
\hspace{1em}Pythia 1.4B & 0.79 [0.75, 0.81] & 0.21 & -0.00\\
\addlinespace[0.3em]
\multicolumn{4}{l}{\textbf{role (subject/object)}}\\
\hspace{1em}BabyLM 125M & 0.85 [0.80, 0.90] & 0.15 & -0.00\\
\hspace{1em}Pythia 160M & 0.92 [0.88, 0.95] & 0.08 & -0.00\\
\hspace{1em}BabyLM 350M & 0.88 [0.83, 0.93] & 0.12 & -0.00\\
\hspace{1em}Pythia 410M & 0.87 [0.82, 0.91] & 0.13 & -0.00\\
\hspace{1em}BabyLM 1.3B & 0.88 [0.84, 0.92] & 0.12 & -0.00\\
\hspace{1em}Pythia 1.4B & 0.87 [0.82, 0.92] & 0.13 & -0.00\\
\bottomrule
\end{tabular}
\endgroup{}

}

\caption{\label{tbl-residual}How much of the deepest-layer interaction
is attributable to the left context, by model and category.
\emph{Word-built} is the share requiring word-dependent processing, with
its bootstrap interval; \emph{context} is the share an item-independent
carry-over of the context predicts; \emph{null} mismatches items before
the comparison.}

\end{table}%

\bibliography{refs.bib}

\end{document}